\documentclass{article}

\usepackage{microtype}
\usepackage{graphicx}
\usepackage{booktabs} 
\usepackage{colortbl}
\usepackage{multirow}
\usepackage{lipsum}
\usepackage{outlines}
\usepackage{enumitem}
\usepackage{bm}
\usepackage{svg}
\usepackage{tikz}
\usepackage[algo2e,algoruled]{algorithm2e}
\usetikzlibrary{arrows, positioning, shapes.geometric}
\usepackage{xcolor}

\usepackage{subcaption}
\usepackage[pagebackref,breaklinks,colorlinks]{hyperref}        
\hypersetup{
    colorlinks = true,
    linkcolor = BrickRed,
    citecolor = MidnightBlue,
    linktoc=all,
}

\definecolor{mygreen}{rgb}{0.0, 0.5, 0.0} 

\usepackage[accepted]{icml2023}

\usepackage{amsmath}
\usepackage{amssymb}
\usepackage{mathtools}
\usepackage{amsthm}
\usepackage{wrapfig}

\usepackage[subtle]{savetrees}
\usepackage{tabularx}
\newcolumntype{Y}{>{\centering\arraybackslash}X}
\usepackage{array,multirow}

\usepackage[capitalize,noabbrev]{cleveref}

\usepackage{import}

\DeclareMathOperator*{\argmax}{arg\,max}
\DeclareMathOperator*{\argmin}{arg\,min} 

\newcommand{\x}{\mathbf{x}}
\newcommand{\z}{\mathbf{z}}

\newcommand{\cD}{\mathcal{D}}
\newcommand{\cX}{\mathcal{X}}

\newcommand{\cZ}{\mathcal{Z}}
\newcommand{\cL}{\mathcal{L}}

\newcommand{\cG}{\mathcal{G}}
\newcommand{\E}{\mathbb{E}}

\newcommand{\Rm}{\mathbb{R}^m}

\newcommand{\pdata}{p_{\text{data}}}
\newcommand{\ptheta}{p_{\theta}}

\newcommand{\Etheta}{E_{\theta}}

\newcommand{\Lmle}{\cL_{\textit{\tiny{MLE}}}}

\newcommand{\Lc}{\cL_{\textit{\tiny{C}}}}
\newcommand{\Ltot}{\cL_{\textit{\tiny{Tot}}}}

\newcommand{\minus}{\scalebox{0.65}[0.9]{$-$}}

\theoremstyle{plain}

\theoremstyle{definition}

\theoremstyle{remark}

\newcommand{\vs}{\emph{vs.}\xspace}

\newcommand{\eg}{\emph{e.g.}\xspace}
\newcommand{\ie}{\emph{i.e.}\xspace}

\usepackage{pifont}
\newcommand{\cmark}{\ding{51}}
\newcommand{\xmark}{\ding{55}}

\usepackage[textsize=tiny]{todonotes}

\icmltitlerunning{Hybrid Energy Based Model in the Feature Space for Out-of-Distribution Detection}

\usepackage[capitalize]{cleveref}
\crefname{section}{Sec.}{Secs.}
\crefname{table}{Tab.}{Tabs.}
\Crefname{equation}{Eq.}{Eqs.}
\Crefname{figure}{Fig.}{Figs.}

\begin{document}

\twocolumn[
\icmltitle{Hybrid Energy Based Model in the Feature Space\\ for Out-of-Distribution Detection}




\begin{icmlauthorlist}
\icmlauthor{Marc Lafon}{cnam}
\icmlauthor{Elias Ramzi}{cnam,coexya}
\icmlauthor{Clément Rambour}{cnam}
\icmlauthor{Nicolas Thome}{sorbonne}
\end{icmlauthorlist}

\icmlaffiliation{cnam}{Cedric Laboratory, Cnam, Paris, France}
\icmlaffiliation{coexya}{Coexya}
\icmlaffiliation{sorbonne}{Sorbonne Université, CNRS, ISIR, F-75005 Paris, France}

\icmlcorrespondingauthor{Marc Lafon}{marc.lafon@lecnam.net}

\icmlkeywords{Machine Learning, ICML}

\vskip 0.3in
]


\printAffiliationsAndNotice{} 

\begin{abstract}

Out-of-distribution (OOD) detection is a critical requirement for the deployment of deep neural networks. This paper introduces the HEAT model, a new post-hoc OOD detection method estimating the density of in-distribution (ID) samples using hybrid energy-based models (EBM) in the feature space of a pre-trained backbone. HEAT complements prior density estimators of the ID density, \eg~parametric models like the Gaussian Mixture Model (GMM), to provide an accurate yet robust density estimation. A~second contribution is to leverage the EBM framework to provide a unified density estimation and to compose several energy terms. Extensive experiments demonstrate the significance of the two contributions. HEAT sets new state-of-the-art OOD detection results on the CIFAR-10 / CIFAR-100 benchmark as well as on the large-scale Imagenet benchmark. The code is available at: \href{https://github.com/MarcLafon/heatood}{\nolinkurl{github.com/MarcLafon/heatood}}.


\end{abstract}

\section{Introduction}
\label{sec:intro}
Out-of-distribution (OOD) detection is a major safety requirement for the deployment of deep learning models in critical applications, \eg healthcare, autonomous steering, or defense~\cite{BendaleB15, Amodei2016, JanaiGBG20}. Deployed machine learning systems must successfully perform a specific task, \eg image classification, or image segmentation while being able to distinguish \emph{in-distribution} (ID) from OOD samples, in order to abstain from making an arbitrary prediction when facing the latter.

OOD detection is a challenge for state-of-the-art deep neural networks.~Most recent approaches~follow a post-hoc strategy \cite{hendrycks17baseline, Liang2018, Liu2020, Sehwag2021, Sun2022, Wang2022} suitable for real-world purpose, which offers the possibility to leverage state-of-the-art models for the main prediction task and to maintain their performances. It also relaxes the need for very demanding training processes, which can be prohibitive with huge deep neural nets and foundation models~\cite{bommasani2021opportunities,Radford2021, Rombach_2022_CVPR, Alayrac2022}.

\emph{Post-hoc} methods exploit the feature space of a pre-trained network and attempt at estimating the density of ID features to address OOD detection.~Existing ID density estimation methods include Gaussian Mixture Models (GMMs) \cite{mahalanobis2018, Sehwag2021}, the nearest neighbors distribution \cite{Sun2022}, or the distribution derived from the energy logits (EL)~\cite{Liu2020}. However, these approaches tend to detect different types of OOD data: for instance, GMMs' density explicitly decreases when moving away from training data, making them effective for far-OOD\footnote{We denote as far (resp. near) OOD samples with classes that are semantically distant (resp. close) from the ID classes.} detection, while EL~benefits from the classifier training to obtain strong results on near-OOD samples \cite{Wang2022}.

 \begin{figure*}[h]
\centering
\import{}{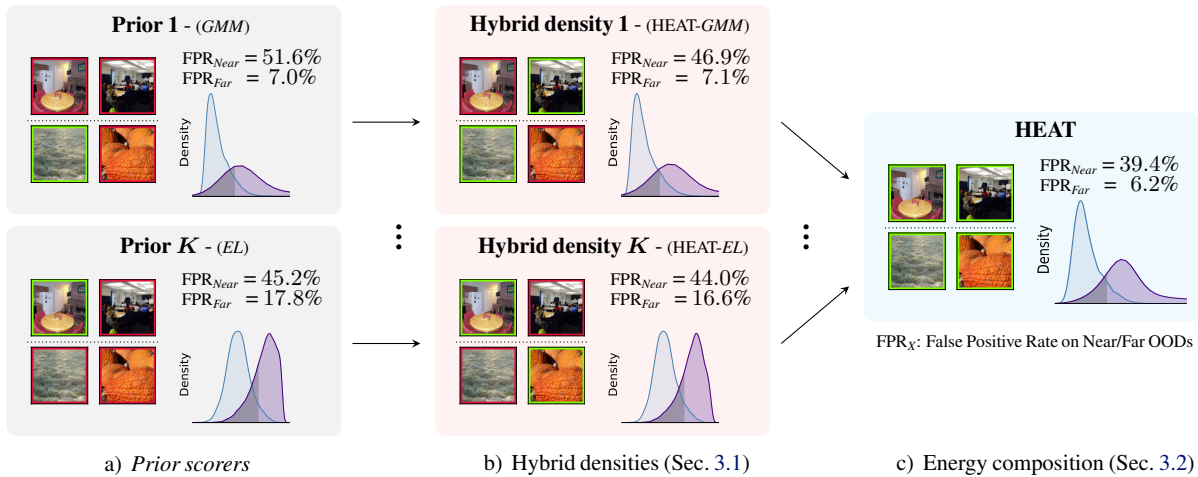}
\caption{\textbf{Illustration of our HEAT model}.
HEAT leverages a) $K$ prior density estimators, such as {\small GMM} or {\small EL}, and overcomes their modeling biases by learning a residual term with an {\small EBM} b) leading to more accurate OOD scorers, \eg~{\small HEAT-GMM} or {\small HEAT-EL}. The second contribution is to combine the different refined scorers using an {\small EBM} energy composition function. The final HEAT prediction c) can thus leverage the strengths of the different OOD scorers, and be effective for both far and near-OOD detection. \vspace{-0.5em}}
\label{fig:intro}
\end{figure*}

In this work, we introduce \textbf{HEAT}, a new density-based OOD detection method which estimate the density of ID samples using a \textbf{H}ybrid \textbf{E}nergy based model in the fe\textbf{AT}ure space of a fixed pre-trained backbone, which provides strong OOD detection performances on both near and far-OOD data. HEAT leverages the energy-based model (EBM) framework \cite{Lecun2006} to build a powerful density estimation method relying on two main components:
\vspace{-0.5em}
\begin{enumerate}[itemsep=0em, topsep=0.5em, parsep=0.5em, leftmargin=*]
    \item \textbf{Energy-based correction} of prior OOD detectors (\eg GMMs or EL) with a data-driven EBM, providing an accurate ID density estimation while benefiting from the strong generalization properties of the priors. The corrected model is carefully trained such that the prior and residual terms achieve optimal cooperation. 
    \item \textbf{Hybrid density estimation} grounded by a~sound energy functions composition combining several sources to improve OOD detection. The energy composition requires a single hyper-parameter, and involves no computational overhead since it is applied at a single layer of the network.
\end{enumerate}

We illustrate {HEAT} in~\cref{fig:intro} using two prior OOD detectors from the literature: SSD+ which is based on GMMs~\cite{Sehwag2021} and EL \cite{Liu2020}, with CIFAR-10 dataset as ID dataset and with six OOD datasets, see ~\cref{sec:exp}.
~We can see in~\cref{fig:intro} that GMM is able to correctly detect far-OOD samples while struggling on near-OOD samples when EL exhibits the opposite behavior. The energy-correction step enhance both priors, reducing the false positive rate (FPR) by -4.7~pts on near-OOD while being stable on far-OOD for GMM, and by -3.2~pts on near-OOD and -1.2~pts for EL. Finally, the energy-composition step produces a hybrid density estimator leading to a better ID density estimation which further improves the OOD detection performances, both for near and far OOD regimes. 

We conduct an extensive experimental validation in~\cref{sec:exp}, showing the importance of our two contributions. HEAT sets new state-of-the-art OOD detection results with CIFAR-10/-100 as ID data, but also on the large-scale Imagenet dataset. HEAT is also agnostic to the prediction backbone (ResNet, ViT) and remains effective in low-data regimes.

\section{Related work}
\label{sec:relworks}

Seminal attempts for OOD detection used supervised methods based on external OOD samples~\cite{Lee2018,malinin2018} or
~``Outlier Exposure'' (OE) \cite{Hendrycks2019} enforcing a uniform OOD distribution. Although OOD datasets can improve OOD detection, their relevance is questionable since collecting representative OOD datasets is arguably impossible as OOD~lie anywhere outside the training distribution \cite{charpentier2020}. It can also have the undesirable effect of learning detectors biased towards certain types of OOD \cite{Wang2022}. 

\textbf{Density-based OOD detection.} Estimating the density of ID training samples to perform OOD detection is a natural strategy that has been widely explored. In their seminal work, \cite{mahalanobis2018} first proposed to approximate the ID features density with a class-conditional {\small GMM}. Subsequent works adopted the same approach by adding slight modifications. For instance, \cite{Sehwag2021} proposed to learn the {\small GMM} density of normalized features without having access to class labels. Recently, \cite{Sun2022} challenged the {\small GMM} distributional assumption by showing that using a deep nearest neighbors approach on normalized features has strong OOD detection performances.

\textbf{E}nergy-\textbf{B}ased \textbf{M}odels (\textbf{EBM}) are another approach to estimate the ID density which have made incredible progress in generative modeling for images in recent years \cite{Xie2016, Du2019, Grathwohl2019}. However, their performances for OOD detection are not yet comparable with OOD methods based on the feature space \cite{Elflein2021}. \cite{Liu2020} have proposed to perform OOD detection with an energy score defined by the \textit{logsumexp} of the logits (EL) of the pre-trained classifier showing improvement over using the classifier's predicted probabilities \cite{hendrycks17baseline}. Furthermore, the authors of EL propose to fine-tune the logits of the classifier using external OOD datasets. Contrarily, we do not use any OOD to learn HEAT but rely on proper EBM training to estimate the ID features density.

\textbf{Energy-based correction.} Our method rely on energy-based correction of a reference model. This idea has been explored in noise contrastive estimation (NCE) \cite{Gutmann2012} where the correction is obtained by discriminative learning. Learning an EBM in cooperation with a generator model has been introduce in \cite{Xie2018} where an EBM learns to refine generated samples and has also been applied to cooperative learning of an EBM with a conditional generator \cite{XieZFZW22}, a VAE \cite{Pang2020, XieZL21, Xiao2021} a normalizing flow \cite{Nijkamp2022, XieZLL22}. Contrarily to our method which is designed for OOD detection, previous works focus on generation and cannot benefit from a fixed prior OOD detector as they use a cooperative learning strategy.

\textbf{Residual learning.}
Training hybrid models, where a  data-driven \emph{residual} complements an approximate predictor, has been proposed in several context, \eg in complex dynamic forecasting~\cite{leguen2021augmenting}, in  NLP~\cite{Bakhtin2021}, in video prediction \cite{Guen_2020_CVPR, leguen22combo}, or in robotics
~\cite{Zeng2020TossingBotLT}. Such residual approaches have also emerged for OOD detection. ResFlow ~\cite{Zisselman2020} uses a normalizing flow (NF) to learn the residual of a Gaussian density for OOD detection. The approach is related to ours, but NFs require invertible mapping, which intrinsically limit their expressive power and make the learned residual less accurate. Also, ViM~\cite{Wang2022} proposes to model the residual of the ID density by using the complement to a linear manifold on the ID manifold. With HEAT, we can learn a non-linear residual and include a residual from different energy terms to improve ID density modeling. We verify experimentally that HEAT significantly outperforms these two baselines for OOD detection.

\textbf{Ensembling \& composition.}
The question of merging several networks, also known as \emph{ensembling} \cite{deepensembles2017} has been among the first and most successful approaches for OOD detection.~The ensemble can include different backbones or different training variants.~For OOD detection, several post-hoc approaches also model the ID density at different layer depth of a pre-trained model, the overall density score being obtained by ensembling such predictions \cite{mahalanobis2018, Sastry2020, Zisselman2020}. The main limitation of these approaches relates to their computational cost since the inference time is proportional to the number of networks. The overhead quickly becomes prohibitive in contexts with limited resources. Several sources of prior densities are combined in \cite{Wang2022} to refine OOD detection. Our approach is a general  framework adapting the EBM composition model \cite{Du2020, Du2021} to OOD detection, and can thus include several hybrid energy terms to refine ID density estimation. In terms of computational cost, we apply our model at a single layer of the network, bringing essentially no computational overhead at inference (see \cref{sec:sup_model_analysis}).

\section{HEAT for OOD detection}
\label{sec:method}
\begin{figure*}[h]
\centering
\import{}{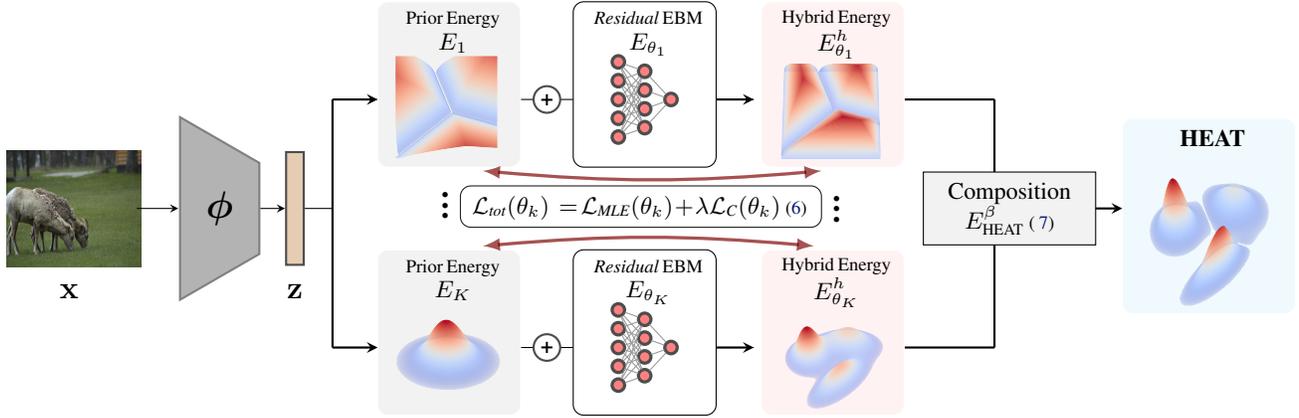}
\vspace{-2em}
\caption{
\textbf{Schematic view of the HEAT model for OOD detection}. Each selected prior density estimator $q_k$ is expressed as an {\small EBM}, $q_k(\bm{z})\propto\exp{(\minus E_{q_k}(\bm{z}))}$, and is refined with its own residual {\small EBM} parameterized with a neural network: The energy for each prior $E_k$ (\eg~{\small EL}, {\small GMM}) is corrected by a residual energy $E_{\theta_k}$ to produce an hybrid energy $E^{h}_{\theta_k}$ (cf.~\cref{sec:refining_scorer}).
Then all hybrid energies are composed to produce HEAT's energy $E_{\textit{HEAT}}^\beta$ (cf.~\cref{sec:combining_energy}), which is used as uncertainty score for OOD detection.
}
\label{fig:model}
\end{figure*}

In this section, we describe the proposed HEAT model to estimate the density of in-distribution (ID) features using a hybrid energy-based model (EBM). We remind that we place ourselves in the difficult but realistic case where only ID samples are available, and we do not use any OOD samples for density estimation. Also, HEAT is a post-hoc approach estimating the density of the latent space of a pre-trained prediction model, as in~\cite{mahalanobis2018,Wang2022,Sehwag2021,Sun2022}.

Let $p(\x)$ be the probability of ID samples, where $\x \in \cX$, and $\z=\phi(\x) \in \cZ$ denotes the  network's embedding of $\x$ with ~$\cZ$ the latent space at the penultimate layer of a pre-trained prediction model $f$, \eg a deep neural net for classification. We aim at estimating $p(\z | \cD)$ with ${\cD:=\{\x_i\}_{i=1}^N}$ the ID training dataset\footnote{we ignore the dependence to $\cD$ in the following and denote the sought density as $p(\z)$.}.

We illustrate the two main components at the core of HEAT in \cref{fig:model}.
Firstly, we introduce a hybrid density estimation to refine a set of prior densities $\{q_k(z)\}_{1\leq k \leq K}$ by complementing each of them with a residual EBM. 
Secondly, we propose to compose several hybrid density estimations based on different priors, which capture different facets of ID density distributions.

\subsection{Hybrid Energy-based density estimation}\label{sec:refining_scorer}

The main motivation in hybrid EBM density estimation is to leverage existing models that rely on specific assumptions on the form of the density $p(\z)$, \eg {\small EL}~\cite{Liu2020}, which captures class-specific information in the logit vector, or SSD~\cite{Sehwag2021} which uses a {\small GMM}. These approaches have appealing properties: {\small GMM} is a parametric model relying on few parameters thus exhibiting strong generalization performances, and {\small EL} benefits from classification training. However, their underlying modeling assumptions intrinsically limit their expressiveness which leads to coarse boundaries between ID and OOD, and they generally fail at discriminating between ambiguous data.

\paragraph{Hybrid EBM model.} Formally, let $q_k(\z)$ be a density estimator inducing an OOD-prior among a set of K priors $\{q_k(\z)\}_{1\leq k \leq K}$. 
~We propose to refine its estimated density by learning a residual model $p_{\theta_k}^r(\z)$, such that our hybrid density estimation is performed by $p_{\theta_k}^h(\z)$ as follows: 
\begin{align}
\label{eq:energy_correction}
p_{\theta_k}^h(\z) = \frac{1}{Z(\theta_k)} p_{\theta_k}^r(\z)q_k(\z),
\end{align}
with $Z(\theta_k) = \int p_{\theta_k}^r(\z)q_k(\z) \mathrm{d}\z$ the normalization constant. We propose to learn the residual density $p_{\theta_k}^r(\z)$ with an EBM: $p_{\theta_k}^r(\z) \propto \exp{(\minus E_{\theta_k}(\z))}$. From \cref{eq:energy_correction}, we can derive a hybrid energy ${E^h_{\theta_k}(\z) = E_{q_k}(\z)+E_{\theta_k}(\z)}$ and express $p_{\theta_k}^h(\z)$ as follows: 
\begin{align}\label{eq:energy_hybrid}
p_{\theta_k}^h(\z) = \frac{1}{Z(\theta_k)} \exp{\left(\minus E^h_{\theta_k}(\z)\right)},
\end{align}
with $E_{q_k} = \minus\log{q_k}(\z)$ the energy from the prior. The goal of the residual energy $E_{\theta_k}(\z)$ is to compensate for the lack of accuracy of the energy of the prior density $q_k(\z)$. We~choose to parameterize it with a neural network, as shown in \cref{fig:model}. This gives our EBM density estimation the required expressive power to approximate the residual term.

\paragraph{Hybrid EBM training.} The hybrid model energy $E^h_{\theta_k}(\z)$ can be learned via maximum likelihood estimation (MLE), which amounts to perform stochastic
gradient descent with the following loss (cf. \cref{sec:sup_ebm} for details):
\begin{equation}
\label{eq:CD}
\Lmle(\theta_k) =\E_{\z \sim p_{in} }\big[E_{\theta_k}(\z)\big] - \E_{\z' \sim p_{\theta_k}^h}\big[E_{\theta_k}(\z')\big],
\end{equation}
with $\z \sim p_{in}$ being the true distribution of the features from the dataset. Minimizing \cref{eq:CD} has for effect to lower the energy of real samples while raising the energy of generated ones.
To learn a residual model we \emph{must} sample $\z'$ from the hybrid model $p_{\theta_k}^h$. To do so, we follow previous works on {\small EBM} training \cite{Du2019} and exploit stochastic gradient Langevin dynamics ({\small SGLD})~\cite{WellingT11}. {\small SGLD} sampling consists in gradient descent on the energy function:
\begin{equation}
\label{eq:SGLD_basic}
\z_{t+1} = \z_{t} - \frac{\eta}{2} \nabla_{\z} E_{\theta_k}^h(\z_{t}) + \sqrt{\eta}~w_t, \;\text{~with $w_t\sim\mathcal N(\bm 0, \bm I)$}
\end{equation}
where $\eta$ is the step size,~the chain being initiated with $\z_0 \sim q_k$. The residual energy corrects the prior density by raising (resp. lowering) the energies in areas where the prior over- (resp. under-) estimates $p_{in}$. It then does so for the current hybrid model $E_{\theta_k}^h$. The overall training hybrid {\small EBM} scheme is summarized \cref{alg:HEAT_training}.

\textbf{Controlling the residual.}\label{sec:regul} As our goal is to learn a residual model over $q$, we must prevent the energy-correction term $E_{\theta_k}$ to take too large values thus canceling the benefit from the prior model $q_k$. Therefore, we introduce an additional loss term preventing the hybrid model from deviating to much from the prior density:
\begin{align}\label{eq:regul}
\Lc(\theta_k) =\E_{p_{in}, p_{\theta_k}^h}\big[~(E^h_{\theta_k} - E_{q_k})^2~\big].
\end{align}

The final loss is then:
\begin{align}\label{eq:heat_training_objective}
\Ltot(\theta_k) =\Lmle(\theta_k) + \lambda\ \Lc(\theta_k),
\end{align}
where $\lambda$ is an hyper-parameter balancing between the two losses. Although $\Lc(\theta_k)$ in \cref{eq:regul} rewrites as $\E_{p_{in}, p_{\theta_k}^h}\big[E_{\theta_k}^2\big]$, we point out that its objective goes beyond a standard $\ell_2$-regularization used to stabilize training. It has the more fundamental role of balancing the prior and the residual energy terms in order to drive a proper cooperation.  

\begin{algorithm}[!h]
\caption{Hybrid Energy Based Model Training}\label{alg:HEAT_training}
\SetKwInOut{Input}{input} \SetKwInOut{Output}{output} 
\Input{
    ~Features $\mathcal D_z$, 
    ID-Prior $(q_k,E_{q_k})$, $\lambda$, $\alpha$ and $\eta$.\vspace{.75ex}
}
\Output{
Hybrid EBM $E_{\theta_k}^h = E_{q_k}(\bm z) + E_{\theta_k}(\bm z)$. ~~\small{\textcolor{mygreen}{// cf. Eq. (\ref{eq:energy_hybrid})}}
}
 \While{not converged}{\vspace{.75ex}
  Sample $\bm z \in \mathcal D_z$ and $\bm z_0' \sim q_k$ \vspace{.75ex}
  
  \For{$0 \leq t \leq T-1$}  { 
    $w \sim \mathcal N(\bm 0,\bm I)$
    
    $\bm z_{t+1}' \leftarrow \bm z_{t'} - \frac{\eta}{2} \nabla_{\bm z} E_{\theta_k}^h(\bm z_{t}') + \sqrt{\eta} w$ ~~\small{\textcolor{mygreen}{// SGLD, \cref{eq:SGLD_basic}}}
  }

  

  $\mathcal L_{Tot}(\theta_k) = \mathcal{L}_{MLE}(\theta_k) +  \lambda \mathcal L_c(\theta_k)$ ~~\small{\textcolor{mygreen}{// cf. Eq. (\ref{eq:heat_training_objective})}}\vspace{.75ex}
  
  $\theta_k \leftarrow \theta_k -\alpha \nabla_{\theta_k} \mathcal L_{Tot}(\theta_k)$\vspace{.75ex}

  
 }
\end{algorithm}

\subsection{Composition of refined prior density estimators}
\label{sec:combining_energy}

In this section, we motivate the choice of prior OOD scorers that we correct, and how to efficiently compose them within our HEAT framework.  

\paragraph{Selected OOD-Priors.}\label{sec:select_ood_priors} As previously stated, {\small EL} and {\small{GMM}} show complementary OOD detection performances, {\small EL} being useful to discriminate class ambiguities while {\small{GMM}} is effective on far-OOD. 
Additionally they can be directly interpreted as energy-based models and thus can  easily be refined and composed with HEAT. Based on the energy from the logits derived in~\cite{Liu2020} we express the hybrid energy {\small{HEAT-EL}} as ${E_{\theta_l}^h(\z) = \minus  \log \sum_c e^{ f(\z)[c]} + E_{\theta_l}^r(\z)}$ where $f(.)[c]$ denotes the logit associated to the class $c$. 

For the {\small GMM} prior, we derive an energy from the Mahalanobis distances to each class centroid. Giving the following expression for our hybrid {\small{HEAT-GMM}}'s energy $ {E_{\theta_g}^h(\z) = - \log \sum_c e^{\minus \frac{1}{2}(\z - \mu_c)^T\bm \Sigma^{-1} (\z - \mu_c)} + E_{\theta_g}^r(\z)}$ with $\Sigma$ and $\mu_c$ being the empirical covariance matrix and mean feature for class $c$.  {\small{HEAT-GMM}}'s energy is computed on the $\z$ vector in \cref{fig:model}, which is obtained by average pooling from the preceding tensor in the network.  
To improve HEAT's OOD detection performances, we propose to further exploit feature volume prior to the pooling operation (\eg average pooling) as we hypothesize that it contains more information relevant to OOD detection. To do so, we compute the vector of second-order moments of the feature volume by using a std-pooling operator and subsequently model the density of the second-order features with a {\small{GMM}}. This leads to a third hybrid {\small EBM} denoted as {\small{HEAT-GMM}$_{std}$}. 

Note that our HEAT method can be extended to $K$ prior scorers, provided that they can write as an {\small EBM} and that they are differentiable in order to perform {\small SGLD} sampling. Interesting extensions would include adapting the approach to other state-of-the-art OOD detectors, such as a soft-{\small KNN}~\cite{Sun2022} or ViM~\cite{Wang2022}. We leave these non-trivial extensions for future works.

\paragraph{Composition strategy.}
 The {\small EBM} framework offers a principled way to make a composition \cite{Du2020} of energy functions. Given $K$ corrected energy functions $E_{\theta_k}^h$, such that: $p_{\theta_k}^h \propto \exp(\minus (E_{\theta_k}(\z) + E_{q_k}(\z)) $, we introduce the following composition function:
\begin{equation}
\label{eq:fusion_function}
E_{\text{HEAT}}^{\beta} = \frac{1}{\beta} \log{\sum_{k=1}^K e^{\beta E_{\theta_k}^h}}
\end{equation}
Depending on $\beta$, $E^\beta_{\text{HEAT}}$ can recover a sum of energies ($\beta=0$), \ie a product of probabilities. For $\beta=\minus 1$, $E^\beta_{\text{HEAT}}$ is equivalent to the \emph{logsumexp} operator, \ie a sum of probabilities. \textcolor{black}{More details in~\cref{sec:sup_compo}.} Moreover, unlike previous approaches that require learning a set of weights \cite{mahalanobis2018, Zisselman2020}, HEAT's composition only requires tuning a single hyper-parameter, \ie $\beta$ which has a clear interpretation.

The composition strategy adopted in HEAT is also scalable since: i) we work in the feature space $\z=\phi(\x) \in \cZ$ of controlled dimension (\eg $1024$ even for the CLIP foundation model~\cite{Radford2021}), and ii) our energy-based correction uses a relatively small model (we use a $6$-layers MLP in practice). We study the computational cost of HEAT in~\cref{sec:sup_model_analysis} and show the large gain in efficiency compared to \eg deep ensembles.

\textbf{OOD detection with HEAT.}
Finally, we use the learned and composed energy of HEAT, $E_{\text{HEAT}}^{\beta}$ in \cref{eq:fusion_function}, as an uncertainty score to detect OOD samples.

\section{Experiments}
\label{sec:exp}
\textbf{Datasets.} We validate HEAT on several benchmarks. The two commonly used CIFAR-$10$ and CIFAR-$100$~\cite{krizhevsky2009learning} benchmarks as in~\cite{Sehwag2021,Sun2022}. We also conduct experiments on the large-scale Imagenet~\cite{deng2009imagenet} dataset. More details in~\cref{sec:sup_exp}.

\textbf{Evaluation metrics.} We report the following standard metrics used in the literature \cite{hendrycks17baseline}: the area under the receiver operating characteristic curve ({\small AUC}) and the false positive rate at a threshold corresponding to a true positive rate of 95\% ({\small FPR95}).

\textbf{Implementation details.} All results on CIFAR-10 and CIFAR-100 are reported using a ResNet-34~\cite{resnet2015}, on Imagenet we use the pre-trained ResNet-50 from \texttt{PyTorch}~\cite{NEURIPS2019_9015}. We detail all implementation details in~\cref{sec:sup_exp}.

\textbf{Baselines.} We perform extensive validation of HEAT \vs several recent state-of-the-art baselines, including the maximum softmax probability ({\small MSP}) \cite{hendrycks17baseline}, {\small ODIN} \cite{odin2018}, Energy-logits \cite{Liu2020}, SSD \cite{Sehwag2021}, {\small KNN} \cite{Sun2022} and {\small ViM}~\cite{Wang2022}. We apply our energy-based correction of {\small{EL}}, {\small{GMM}} and {\small{GMM}$_{std}$} that we then denote as {\small{HEAT-EL}}, {\small{HEAT-GMM}} and {\small{HEAT-GMM}$_{std}$}. We choose those priors as they can naturally be written as energy models as described in~\cref{sec:refining_scorer}, furthermore, they are strong baselines and combining them allows us to take advantage of their respective strengths (discussed in~\cref{sec:select_ood_priors}). All the baselines are compared using the same backbone trained with the standard cross-entropy loss.

\begin{table*}[!ht]
\caption{Refinement of Energy-logits~\cite{Liu2020}~({\small{EL}}) and {\small{GMM}}, {\small{GMM}} with std-pooling ({\small{GMM}}$_{std}$) with our energy-based correction on CIFAR-10 and CIFAR-100 as in-distribution datasets. Results are reported with \text{{\small FPR95}$\downarrow$ / {\small AUC} $\uparrow$}.}
\centering
\setlength\tabcolsep{10pt}
\resizebox{0.9\linewidth}{!}{%
\begin{tabular}{l l cccccc|c }
\toprule
    & \multirow{2}{*}{\textbf{Method}} & \multicolumn{2}{c}{\textit{Near-OOD}} & \multicolumn{2}{c}{\textit{Mid-OOD}} & \multicolumn{2}{c|}{\textit{Far-OOD}} & \multirow{2}{*}{\textbf{Average}}\\
    &  &\textbf{C-100/10} & \textbf{TinyIN} & \textbf{LSUN} & \textbf{Places} & \textbf{Textures} & \textbf{SVHN} & \\
    \toprule
    \multirow{6}{*}{\rotatebox[origin=c]{90}{\textbf{CIFAR-10}}}
    & EL & {48.4} / 86.9 & {41.9} / 88.2 & {33.7} / 92.6 & {35.7} / 91.0 &  30.7 / 92.9 & 4.9 / 99.0 &  {32.6} / 91.8  \\
    & HEAT-EL & \textbf{47.3} / \textbf{88.0}&  \textbf{40.7} / \textbf{88.9}&  \textbf{30.8} / \textbf{93.4}&  \textbf{33.8} / \textbf{91.8}&  \textbf{28.8} / \textbf{93.9}&  \textbf{4.5} / \textbf{99.1}&   \textbf{31.0} / \textbf{92.5} \\
    \cmidrule {2-9}
    & GMM & 52.6 / 89.0 & 50.9 / 89.5 &  47.1 / 92.4 & 46.4 / 91.2 &  \textbf{13.1 / 97.8} & \textbf{0.9 / 99.8} & 35.1 / {93.3} \\
    & HEAT-GMM& \textbf{49.0} / \textbf{89.8} & \textbf{44.8} / \textbf{90.4} & \textbf{40.5} / \textbf{93.2} & \textbf{40.4 / 92.0} & {13.4 / 97.7}&   \textbf{0.8 / 99.8} &   \textbf{31.5 / 93.8} \\
    \cmidrule {2-9}
    & GMM$_{std}$ & 58.4 / 84.9&   50.6 / 87.9&   {32.2 / 94.5}&   {38.5} / 91.8&   13.8 / \textbf{97.6} & \textbf{2.5 / 99.5} &  {32.7} / 92.7 \\
    & HEAT-GMM$_{std}$& \textbf{56.1 / 86.1} &  \textbf{47.8 / 88.7} &  \textbf{28.2 / 95.2} &  \textbf{35.8 / 92.5} &  \textbf{13.3} / \textbf{97.5} &  2.7 / 99.4&   \textbf{30.7 / 93.2} \\
    \midrule
    \midrule
    \multirow{6}{*}{\rotatebox[origin=c]{90}{\textbf{CIFAR-100}}}
    & EL & {80.6} / {76.9} & 79.4 / 76.5 & 87.6 / 71.7 & 83.1 / 74.7 & 62.4 / 85.2 & 53.0 / 88.9 &  74.3 / 79.0  \\
    & HEAT-EL & \textbf{80.1 / 77.2}&  \textbf{77.6 / 77.5}&  \textbf{87.2 / 72.2}&  \textbf{81.8 / 75.0}&  \textbf{61.5 / 85.8}&  \textbf{47.5 / 90.2}&   \textbf{72.6 / 79.6} \\
    \cmidrule {2-9}
    & GMM & 85.6 / 73.6&  82.5 / 77.2&  87.8 / 73.7&  84.5 / 74.4&  \textbf{36.7} / \textbf{92.4}&  20.0 / 96.3&   66.2 / 81.3 \\
    & HEAT-GMM & \textbf{84.2 / 74.8} & \textbf{80.5 / 78.5} & \textbf{86.4 / 74.8} & \textbf{82.7 / 75.9} & 37.9 / \textbf{92.2}&  \textbf{17.8 / 96.7} & \textbf{64.9 / 82.1} \\
    \cmidrule {2-9}
    & GMM$_{std}$ & 91.4 / 67.9&   84.3 / 74.8&   83.4 / 75.2&   83.5 / 75.2 & \textbf{40.6 / 91.3} & 36.7 / 93.1&   70.0 / 79.6 \\
    & HEAT-GMM$_{std}$ & \textbf{89.1 / 70.3} & \textbf{82.2 / 76.2} & \textbf{82.3 / 76.1} & \textbf{81.4 / 76.7} & 42.9 / 90.7&  \textbf{32.9 / 93.8} & \textbf{68.5 / 80.6} \\
\bottomrule
\end{tabular}
}
\label{tab:ablation_prior_scorer}
\end{table*}

\subsection{HEAT improvements}
\label{sec:ablation_study}

In this section we study the different components of HEAT. In~\cref{tab:ablation_prior_scorer} we show that learning a residual correction term with HEAT improves the OOD detection performances of prior scorers. In~\cref{tab:ablation_ebm_vs_heat} we show the interest of learning a residual model as described in~\cref{sec:refining_scorer} rather than a standard fully data-driven energy-based model. Finally in~\cref{tab:ablation_fusion} we show how using the energy composition improves OOD detection.

\paragraph{Correcting prior scorers.} In~\cref{tab:ablation_prior_scorer} we demonstrate the effectiveness of energy-based correction to improve different prior OOD scorers on two ID dataset: CIFAR-10 and CIFAR-100. We show that across the two ID datasets and for all prior scorers, using a residual corrections always improves the aggregated results, \eg for {\small GMM} -3.6 pts {\small FPR95} on CIFAR-10 and -1.3 pts {\small FPR95} on CIFAR-100. Furthermore on near-OOD and mid-OOD learning our correction always improves the prior scores, \eg on LSUN with CIFAR-10 as ID dataset the correction improves {\small EL} by -2.9 pts {\small FPR95}, {\small GMM} by -6.6 pts {\small FPR95} and -4 pts {\small FPR95} for {\small GMM}$_{std}$. On far-OOD the corrected scorers performs at least on par with the base scorers, and can further improve it, \eg on SVHN when CIFAR-100 is the ID datasets, the correction improves by -5.5pts {\small FPR95}, -2.2 pts {\small FPR95} and -3.8pts {\small FPR95}, {\small EL}, {\small GMM}, {\small GMM}$_{std}$ respectively. Overall~\cref{tab:ablation_prior_scorer} clearly validates the relevance of correcting the modeling assumptions of prior scorers with our learned energy-based residual.

\begin{table}[h]
\caption{Comparison of learning a residual model, \ie~{\small{HEAT-GMM}}, \vs learning an {\small{EBM}} and {\small{GMM}}. Results reported with \text{{\small AUC} $\uparrow$}.} 
\centering
\setlength\tabcolsep{2pt}
\resizebox{\linewidth}{!}{%
\begin{tabular}{ll  cccccc|c }
\toprule
    & \multirow{2}{*}{\small{\textbf{Method}}} & \multicolumn{2}{c}{\scriptsize{\textit{Near-OOD}}} & \multicolumn{2}{c}{\scriptsize{\textit{Mid-OOD}}} & \multicolumn{2}{c|}{\scriptsize{\textit{Far-OOD}}} & \multirow{2}{*}{\scriptsize{\textbf{Average}}}\\
    &  & \textbf{\small{C-100/10}} & \small{\textbf{TinyIN}} & \small{\textbf{LSUN}} & \small{\textbf{Places}} & \small{\textbf{Textures}} & \small{\textbf{SVHN}} & \\
    \toprule
    \multirow{3}{*}{\rotatebox[origin=c]{90}{{\textbf{C-10}}}} & \small{GMM} & 89.0 & 89.5 &  92.4 & 91.2 &  \textbf{97.7}& \textbf{99.8} &  {93.3} \\
    & \small{EBM} & 89.4&   89.9&   \textbf{93.8} & 91.8& 96.2&  99.0&  93.3\\
    & \small{HEAT-GMM} & \textbf{89.8} & \textbf{90.4} & {93.2} &\textbf{92.0} & \textbf{97.7}&   \textbf{99.8} &  \textbf{93.8} \\
    \midrule
    \midrule
    \multirow{3}{*}{\rotatebox[origin=c]{90}{{\textbf{C-100}}}}
     & \small{GMM} &  73.6 & 77.0  & 73.8  & 74.5  & \textbf{92.4}  & {96.4}  &  {81.3} \\
    & \small{EBM} & \textbf{74.8} & \textbf{ 79.7}&  71.9&  75.4&  84.5&  91.0&  79.5 \\
    & \small{HEAT-GMM} & \textbf{74.8} & {78.5} & \textbf{74.8} & \textbf{75.9} & \textbf{92.2}&  \textbf{96.7} & \textbf{82.1} \\
\bottomrule
\end{tabular}
}
\label{tab:ablation_ebm_vs_heat}
\end{table}

\paragraph{Learning a residual model.} In~\cref{tab:ablation_ebm_vs_heat} we compare learning an EBM (cf.~\cref{sec:sup_ebm}) \vs our residual training using a GMM prior ({\small{HEAT-GMM}}) of~\cref{sec:method} on CIFAR-10 and CIFAR-100. The {\small EBM} is a fully data-driven approach, which learns the density of ID samples without any prior distribution model. On both datasets, our residual training leads to better performances than the EBM, \eg +2.6 pts {\small AUC} on CIFAR-100. On near-OOD, both the residual training and the {\small EBM} perform on par. On far-OOD, our residual training takes advantage of the good performances of the prior scorer, \ie {\small GMM}, and significantly outperforms the {\small EBM}, especially on CIFAR-100, with \eg +7.7 pts {\small AUC} on Textures. Our residual training combines the strengths of {\small GMM} and {\small EBM}s: Gaussian modelization by design penalizes samples far away from the training dataset and thus eases far-OOD's detection, whereas {\small EBM} may overfit in this case. On the other hand, near-OOD detection requires a too complex density estimation for simple parametric distribution models such as {\small GMM}s.

\begin{table}[h]
\caption{Aggregated performances on CIFAR-10 and CIFAR-100 for the energy composition of the refined OOD scorers of~\cref{tab:ablation_prior_scorer}.}
\setlength\tabcolsep{5pt}
\centering
\small
\resizebox{0.95\linewidth}{!}{%
\begin{tabular}{c c c | cc | cc}
\toprule
\scriptsize{HEAT} & \scriptsize{HEAT}& \scriptsize{HEAT} & \multicolumn{2}{c}{{\scriptsize{\textbf{CIFAR-10}} }} &\multicolumn{2}{|c}{{\scriptsize{\textbf{CIFAR-100}} }} \\
 \scriptsize{-GMM} & \scriptsize{-GMM$_{std}$}  & \scriptsize{-EL} & \scriptsize{\text{FPR95$\downarrow$}} &\scriptsize{\text{AUC $\uparrow$}}& \scriptsize{\text{FPR95$\downarrow$}} &\scriptsize{\text{AUC $\uparrow$}}\\
\toprule
\textcolor{blue}{\cmark} & \xmark &\xmark & 31.5 & 93.8 & 64.9 & 82.1 \\
\xmark & \textcolor{blue}{\cmark} &\xmark & 30.7 & 93.2 & 68.5 & 80.6 \\
\xmark & \xmark & \textcolor{blue}{\cmark} & 31.0 & 92.5 & 72.6 & 79.6\\
\textcolor{blue}{\cmark} & \textcolor{blue}{\cmark} &\xmark & 25.6 & 94.6 & 64.3 & 82.7 \\
\textcolor{blue}{\cmark} & \xmark & \textcolor{blue}{\cmark} & 28.0 & 94.1 & 65.5 & 82.4 \\
\xmark & \textcolor{blue}{\cmark} & \textcolor{blue}{\cmark} & 23.6 & 94.6 & 66.6 & 82.1\\[1pt]
\rowcolor[gray]{.95} \textcolor{blue}{\cmark} & \textcolor{blue}{\cmark} & \textcolor{blue}{\cmark} & \textbf{23.5} & \textbf{94.8} & \textbf{63.9} & \textbf{83.0}\\
\bottomrule
\end{tabular}
}
\label{tab:ablation_fusion}
\end{table}

\begin{table*}[ht]
\caption{\textbf{Results on CIFAR-10 \& CIFAR-100.} All methods are based on a pre-trained ResNet-34 trained on the ID dataset only. $\uparrow$ indicates larger is better and $\downarrow$ the opposite. Best results are in bold, second best underlined. Results are reported with \text{{\small FPR95}$\downarrow$ / {\small AUC} $\uparrow$}.}
\centering
\setlength\tabcolsep{8pt}
\centering
\resizebox{0.99\linewidth}{!}{%
    \begin{tabular}{p{0.1pt}l  c c  c c  c c  |c}
        \toprule
           &  \multirow{2}{*}{\textbf{Method}} &  \multicolumn{2}{c}{\textit{Near-OOD}} & \multicolumn{2}{c}{\textit{Mid-OOD}}  & \multicolumn{2}{c|}{\textit{Far-OOD}} &  \multirow{2}{*}{\textbf{Average}} \\
         & & \textbf{C-10/C-100} & \textbf{TinyIN} & \textbf{LSUN} & \textbf{Places} & \textbf{Textures} & \textbf{SVHN} & \\
        \midrule
            \multirow{7}{*}{\rotatebox[origin=c]{90}{\textbf{CIFAR-10}}}
            &MSP  \cite{hendrycks17baseline} &  58.0 / 87.9 & 55.9 / 88.2  & 50.5 / 91.9 & 52.7 / 90.2 &  52.3 / 91.7 & 19.7 / 97.0 & 48.2 / 91.2\\ 
            &ODIN \cite{odin2018} & {48.4} / 86.0 & {42.2} / 87.3 & {32.6} / 92.3 & \underline{35.6} / 90.4 & 29.4 / 92.6 & 7.8 / 98.3 &  {32.6} / 91.1\\ 
            &KNN \cite{Sun2022} & \underline{47.9} / \textbf{90.3} & 43.1 / \underline{90.6} &  36.1 / \underline{94.1} & 37.9 / \underline{92.7} &  24.9 / 96.0 & 8.1 / 98.6 & {33.0} / \underline{93.7}\\ %
            &ViM \cite{Wang2022} & 44.8 / 89.2&   40.1 / 89.8&   \underline{32.0} / 93.8&   34.3 / 92.2&   17.9 / 96.4&   \underline{3.6} / \underline{99.2}&   \underline{28.8} / {93.4}\\ %
            &SSD+ \cite{Sehwag2021} &  52.6 / 89.0 & 50.9 / 89.5 &  47.1 / 92.4 & 46.4 / 91.2 &  \underline{13.1} / \textbf{97.8} & \textbf{0.9 / 99.8} & 35.1 / {93.3}\\ %
            &EL \cite{Liu2020} &  {48.4} / 86.9 & \underline{41.9} / 88.2 & {33.7} / 92.6 & \underline{35.7} / 91.0 &  30.7 / 92.9 & 4.9 / 99.0 &  {32.6} / 91.8\\ %
            & DICE \cite{sun2022dice} & 51.0 / 85.7&  44.3 / 87.0&  33.3 / 92.3&  35.6 / 90.5&  29.3 / 92.8&  3.6 / 99.2&   32.8 / 91.3\\
            \cmidrule{2-9}
            \rowcolor[gray]{.95} & \textbf{HEAT (ours)} & \textbf{43.1 / 90.2}&  \textbf{35.7 / 91.3}&  \textbf{22.2 / 95.8}& \textbf{27.4 / 93.9}&  \textbf{11.3 / 97.9}&  \textbf{1.1 / 99.8}&  \textbf{23.5 / 94.8} \\
            \midrule
            \midrule
            \multirow{7}{*}{\rotatebox[origin=c]{90}{\textbf{CIFAR-100}}}
            &MSP \cite{hendrycks17baseline} & \textbf{80.0} / \textbf{76.6} & 78.3 / 77.6 & \textbf{83.5} / 74.7 & 81.0 / 76.4 & 72.1 / 81.0 & 62.0 / 86.4 & 76.1 / 78.8\\ %
            &ODIN  \cite{odin2018} & \underline{81.4} / \underline{76.4} & 78.7 / 76.2 & \underline{86.1} / 72.0 & 82.6 / 74.5 & 62.4 / 85.2 & 80.7 / 80.4 & 78.6 / 77.5 \\ %
            &KNN \cite{Sun2022} & 82.1 / 74.5&  \textbf{76.7} / \textbf{80.2}&  90.1 / 74.4&  83.2 / 75.5&  47.2 / 90.2&  35.6 / 93.6&   69.2 / 81.4 \\ %
            &ViM \cite{Wang2022} & 85.8 / 74.3&  \underline{77.5} / \underline{79.6}&  86.2 / \underline{75.3}&   \textbf{79.8} / \textbf{77.6} & 42.3 / {91.9} &   41.3 / 93.2&   68.8 / {82.0}\\ %
            &SSD+ \cite{Sehwag2021}  & 85.6 / 73.6&  82.5 / 77.2&  87.8 / 73.7&  84.5 / 74.4&  \textbf{36.7} / \underline{92.4}&  \textbf{20.0} / \textbf{96.3}&   {66.2} / 81.3 \\ %
            &EL \cite{Liu2020} & \textbf{80.6} / \textbf{76.9} & 79.4 / 76.5 & 87.6 / 71.7 & 83.1 / 74.7 & 62.4 / 85.2 & 53.0 / 88.9 &  74.3 / 79.0 \\ %
            & DICE \cite{sun2022dice} & 81.2 / 75.8&  82.4 / 74.2&  87.8 / 70.4&  84.5 / 73.1&  63.0 / 83.8&  51.9 / 88.1&   75.2 / 77.6\\ %
            \cmidrule{2-9}
            \rowcolor[gray]{.95} & \textbf{HEAT (ours)} & 83.7 / 75.8&  \underline{77.7} / \underline{79.5}&  \textbf{83.4} / \textbf{76.3}&  \textbf{80.0 / 77.8}&  \underline{37.1} / \textbf{92.7}&  \underline{21.7} / \textbf{96.0}&   \textbf{63.9 / 83.0} \\
        \bottomrule
    \end{tabular}
    }
\label{tab:main_c10_c100}
\end{table*}

\paragraph{Composing energy-based scorers.} In~\cref{tab:ablation_fusion} we show that composing different energy-based scores (see~\cref{sec:combining_energy}), \ie the selected OOD prior scorers with our energy-based correction as described in~\cref{sec:refining_scorer}, improves overall performances on CIFAR-10 and CIFAR-100. For instance composing our {\small{HEAT-GMM}} and {\small{HEAT-GMM}}$_{std}$ leads to improvements of all reported results, \ie on CIFAR-10 -5.1~pts {\small FPR95} and +0.8~pts {\small AUC} and on CIFAR-100 -0.6 pts {\small FPR95} and +0.6 pts {\small AUC}. Composing the three prior scorers leads to the best results, improving over the best single scorer performances by great margins on CIFAR-10 with -7.1 pts {\small FPR95} and +1 {\small AUC} and with smaller margins on CIFAR-100 -0.8~pts {\small FPR95} and +1.1~pts {\small AUC} on CIFAR-100. This shows the interest of composing different scorers as they detect different types of OOD. Note that while the composition has the best performances our correction model ({\small{HEAT-GMM}}) already has competitive performances on CIFAR-10 and better performances on CIFAR-100 than state-of-the-art methods reported in~\cref{tab:main_c10_c100}.

\subsection{Comparison to state-of-the-art}\label{sec:sota_comparison}
In this section, we present the results of HEAT \vs state-of-the-art methods. In~\cref{tab:main_c10_c100} we present our results with CIFAR-10, and CIFAR-100 as ID data, and in~\cref{tab:ood_results_imagenet} we present our results on the large and complex Imagenet dataset.

\paragraph{CIFAR-10 results.} In~\cref{tab:main_c10_c100} we compare HEAT \vs state-of-the-art methods when using CIFAR-10 as the ID dataset. First, we show that HEAT sets a new state-of-the-art on the aggregated results. It outperforms the prior scorers it corrects, \ie  {\small SSD+} by -11.6 pts {\small FPR95} and Energy-logits by -9.1 pts {\small FPR95}. It also outperforms the previous state-of-the-art methods {\small ViM} by -5.3 pts {\small FPR95} and {\small KNN} by +1.1 pts {\small AUC}. Interestingly we can see that HEAT outperforms other methods because it improves OOD detection on near-, mid-, and far-OOD. On near OOD, it outperforms {\small KNN} by -4.6 pts {\small FPR95} on C-100 and Energy-logits by -6.1 pts {\small FPR95} on TinyIN. On mid-OOD detection, it outperforms ViM by -9.8 pts {\small FPR95} on LSUN and Energy-logits by -8.5 pts {\small FPR95}. Finally, on far-OOD, the performances are similar to {\small SSD+} which is by far the best performing method on this regime.

\paragraph{CIFAR-100 results.} In~\cref{tab:main_c10_c100} we compare HEAT \vs state-of-the-art method when using CIFAR-100 as the ID dataset. HEAT outperforms state-of-the-art methods on aggregated results, with -2.3 pts {\small FPR95} and +1.7 pts {\small AUC} \vs {\small SSD+}. HEAT takes advantage of {\small SSD+} on far-OOD and outperforms other methods (except SSD+) by large margins -13.9 pts {\small FPR95} and +2.4 pts {\small AUC} on SVHN \vs the best  non-parametric data-driven density estimation, \ie {\small ~KNN}. Also, HEAT significantly outperforms {\small SSD+} for near-OOD and mid-OOD, \eg -4.8 pts {\small FPR95} on TinyIN or -4.5 pts {\small FPR95} on Places.

\begin{table*}[!ht]
\caption{\textbf{Results on Imagenet.} All methods use an Imagenet pre-trained {ResNet-50}. Results are reported with \text{{\small FPR95}$\downarrow$ / {\small AUC} $\uparrow$}.}
\setlength\tabcolsep{14pt}
\centering
\resizebox{0.99\linewidth}{!}{%
\begin{tabular}{l  c c  c c | c}
    \toprule
    \textbf{Method} & \textbf{iNaturalist} & \textbf{SUN} & \textbf{Places} & \textbf{Textures} & \textbf{Average}\\
    \midrule
    MSP \cite{hendrycks17baseline} & 52.8 / 88.4&   69.1 / 81.6&   72.1 / 80.5&   66.2 / 80.4&   65.1 / 82.7 \\[1pt]
    ODIN \cite{odin2018} & {41.1}  / {92.3} &   {56.4} / {86.8}&   {64.2} / {84.0}&   46.5 / 87.9&    52.1 / {87.8} \\[1pt]
    {ViM} \cite{Wang2022} & 47.4 / {92.3} &  62.3 / 86.4&  68.6 / 83.3&  15.2 / 96.3&   {48.4} / {89.6} \\[1pt]
    KNN \cite{Sun2022} & 60.0 / 86.2&   70.3 / 80.5&   78.6 / 74.8&   \underline{11.1} / \underline{97.4}&   55.0 / 84.7 \\[1pt]
    SSD+ \cite{Sehwag2021} & 50.0 / 90.7&   66.5 / 83.9&   76.5 / 78.7&   \textbf{5.8 / 98.8}&  {{49.7} / 88.0} \\[1pt]
    EL \cite{Liu2020} & 53.7 / 90.6&   58.8 / {86.6} &   66.0 / {84.0} &   52.4 / 86.7&    57.7 / 87.0 \\[1pt]
    DICE \cite{sun2022dice} & \textbf{26.6} / \underline{94.5} &\textbf{36.5 / 90.8} &\textbf{47.9 / 87.5} &32.6 / 90.4 &\underline{35.9} / \underline{90.9} \\[1pt]
    \midrule
    \rowcolor[gray]{.95} \textbf{HEAT (ours)} & \underline{28.1} / \textbf{94.9}&  \underline{44.6} / \textbf{90.7} &  \underline{58.8} / \underline{86.3}&  {\textbf{5.9} / \textbf{98.7}}&   {\textbf{34.4} / \textbf{92.6}} \\[1pt]
    \bottomrule
\end{tabular}
}
\label{tab:ood_results_imagenet}
\end{table*}

\begin{figure*}[ht]
\vspace{12pt}
\begin{minipage}[t]{0.48\linewidth}
\vspace{0pt}
\begin{minipage}[t]{0.49\linewidth}
    \vspace{0pt}
        \centering
    \includegraphics[width=1\linewidth]{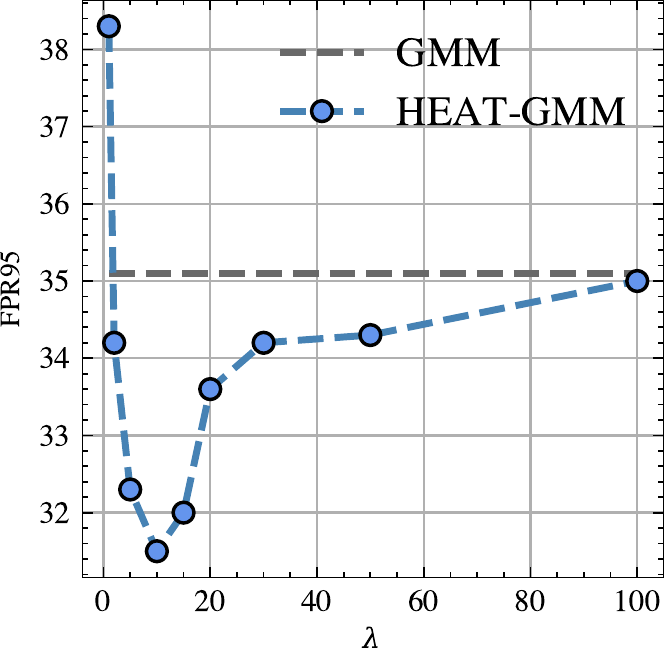}
    \subcaption{$\lambda$ \vs {\small FPR95}$\downarrow$}
    \label{fig:lambda_analysis_auc_c10}
\end{minipage}
\begin{minipage}[t]{0.49\linewidth}
\vspace{0pt}
        \centering
    \includegraphics[width=1\linewidth]{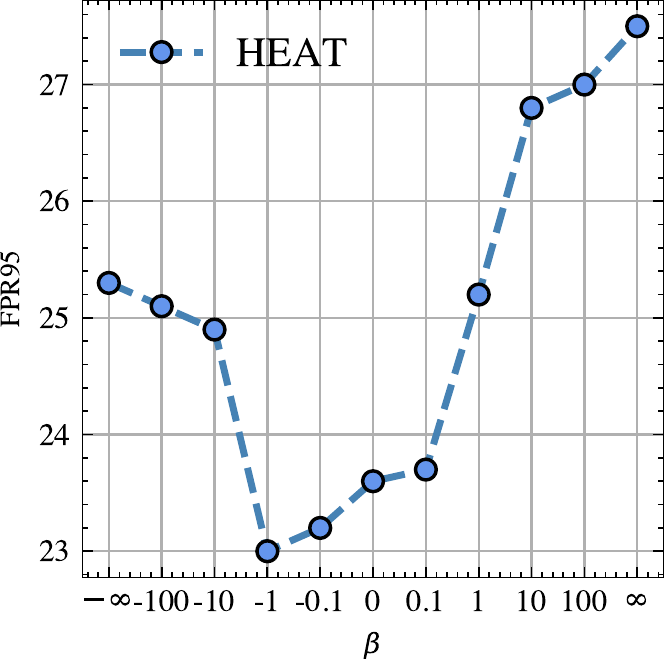}
   \subcaption{$\beta$ \vs {\small FPR95}$\downarrow$}
   \label{fig:beta_analysis_fpr_c10}
\end{minipage}
\caption{On CIFAR-10 ID: (a) impact of $\lambda$ in~\cref{eq:heat_training_objective} \vs ~{\small FPR95} and (b) analysis of $\beta$ in~\cref{eq:fusion_function} \vs ~{\small FPR95}. }
\label{fig:lambda_beta_analysis_c10}
\end{minipage}
\hfill
\begin{minipage}[t]{0.48\linewidth}
\vspace{0pt}
    \centering
    \includegraphics[width=0.93\linewidth]{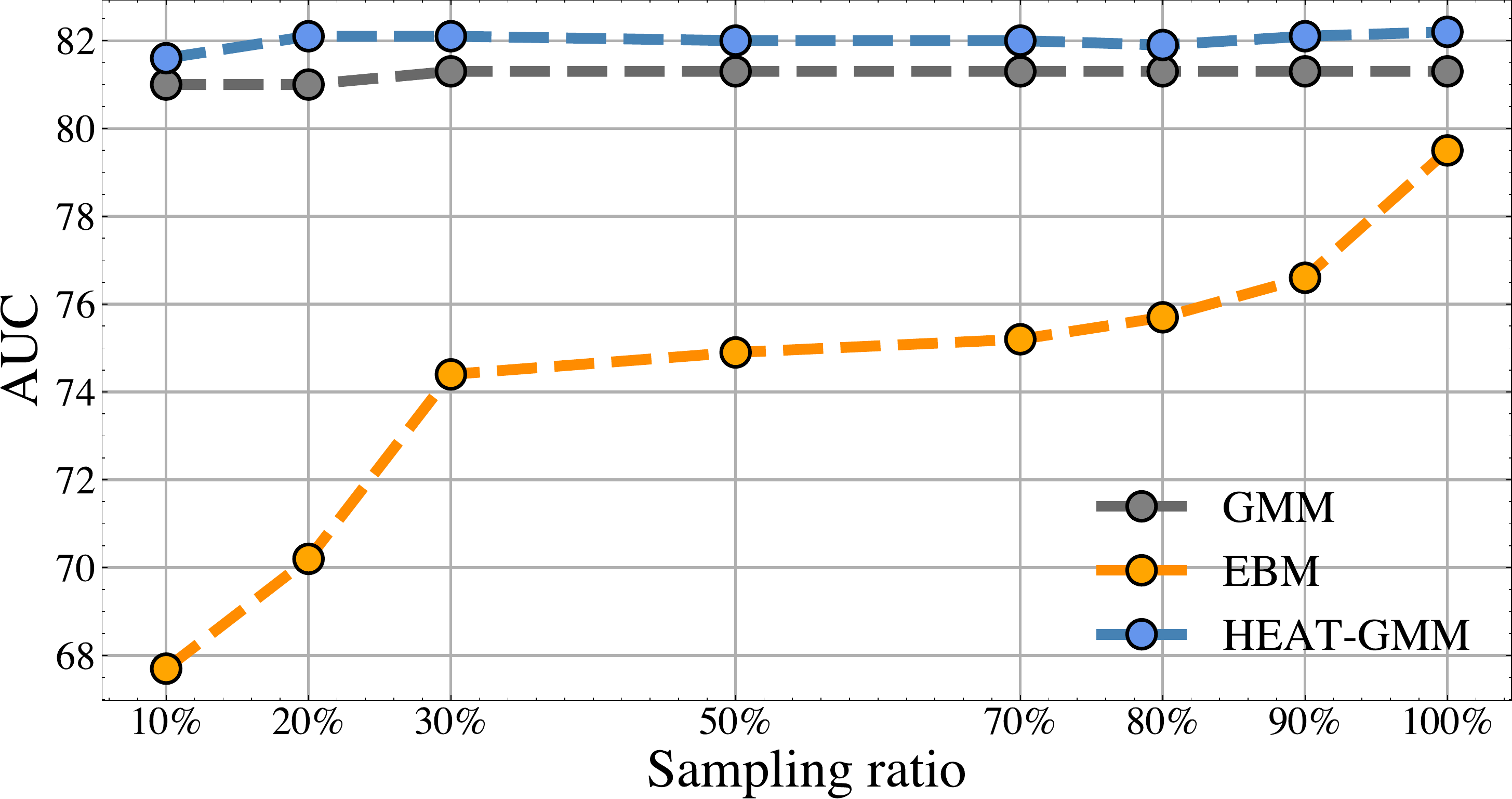}
    \caption{Impact on performances ({\small AUC}$\uparrow$ on CIFAR-100) \vs the number of training data for {\small GMM} density, fully data-driven {\small EBM}, and HEAT. Our hybrid  approach maintains strong performances in low-data regime, in contrast to the fully data-driven {\small EBM}.}
\label{fig:number_training_samples_auc_c100}
\end{minipage}
\end{figure*}

\paragraph{Imagenet results.} In~\cref{tab:ood_results_imagenet} we compare HEAT on the recently introduced~\cite{Sun2022} Imagenet OOD benchmark. 
HEAT sets a new state-of-the-art on this Imagenet benchmark for the aggregated results, with 34.4 {\small FPR95} and 92.6 {\small AUC} which outperforms by -1.5 pts {\small FPR95} and +1.7 pts {\small AUC} \vs the previous best performing method {\small DICE}. Furthermore, HEAT improves the aggregated results because it is a competitive method on each dataset. On far-OOD, \ie Textures, it performs on par with {\small SSD+}, \ie 5.7 {\small FPR95}, the best performing method on this dataset. On mid-OOD, it is the second best method on SUN and on Places behind {\small DICE}. Finally, on near-OOD it performs on par with {\small DICE}. This shows that HEAT can be jointly effective on far-, mid-, and near-OOD detection, whereas state-of-the-art methods are competitive for a specific type of OOD only. For instance the performance of {\small DICE} drops significantly on Textures. {Furthermore, we show in~\cref{sec:sup_dice} that using an energy refined version of DICE instead of EL into HEAT's composition further improve OOD detection results}. This also shows that HEAT performs well on larger scale and more complex datasets such as Imagenet. In~\cref{sec:sup_openood} we show the results of HEAT on the more recent Imagenet OpenOOD benchmarks~\cite{OpenOOD2022}, where we show that HEAT also outperforms state-of-the-art methods. In~\cref{sec:sup_supcon} we show that HEAT also outperforms other methods when using a supervised contrastive backbone. Finally in~\cref{sec:sup_vit} we show that HEAT is also state-of-the-art when using another type of neural network, \ie Vision Transformer~\cite{dosovitskiy2020image}.

\subsection{Model analysis}\label{sec:model_analysis}

In this section we show how HEAT works in a wide range of settings. We show in~\cref{fig:lambda_beta_analysis_c10} the impact of $\lambda$ and $\beta$ and in~\cref{fig:number_training_samples_auc_c100} that HEAT performs well in low data regimes.

\paragraph{Robustness to  $\bm{\lambda}$.} We show in~\cref{fig:lambda_analysis_auc_c10} the impact of $\lambda$ on the {\small FPR95} for CIFAR-10 as the ID dataset. We can observe that for a wide range of $\lambda$, \eg $[2, 50]$, our energy-based correction improves the OOD detection of the prior scorer, \ie {\small GMM}, with ideal values close to $\sim10$. $\lambda$ controls the cooperation between the prior scorer and the learned residual term which can be observed on~\cref{fig:lambda_analysis_auc_c10}. When setting $\lambda$ to a value that is too low there is no control over the energy. The prior density is completely disregarded which will eventually lead to optimization issues resulting in poor detection performances. On the other hand, setting $\lambda$ to a value too high (\eg 100) will constrain the energy too much, resulting in performances closer to that of {\small GMM}. On CIFAR-10 as the ID dataset we observe similar trends in~\cref{sec:sup_model_analysis}.

\textbf{Robustness to $\bm{\beta}$.} We show in~\cref{fig:beta_analysis_fpr_c10} that HEAT is robust wrt. $\beta$ in~\cref{eq:fusion_function}. We remind that $\beta\rightarrow 0$ is equivalent to the mean, $\beta\rightarrow -\infty$ is equivalent to the minimum and $\beta\rightarrow \infty$ is equivalent to the maximum. We show that HEAT is stable to different values of $\beta$, and performs best with values close to 0, this is also the case in~\cref{sec:sup_model_analysis}. Note that we used $\beta=0$ for HEAT in \cref{tab:ood_results_imagenet} and \cref{tab:main_c10_c100} but using a lower value, \ie -1, leads to better results. We hypothesize that using a more advanced $\bm{\beta}$ selection methods could further improve performances.

\begin{figure*}[t]
    \centering
    \includegraphics[width=\linewidth]{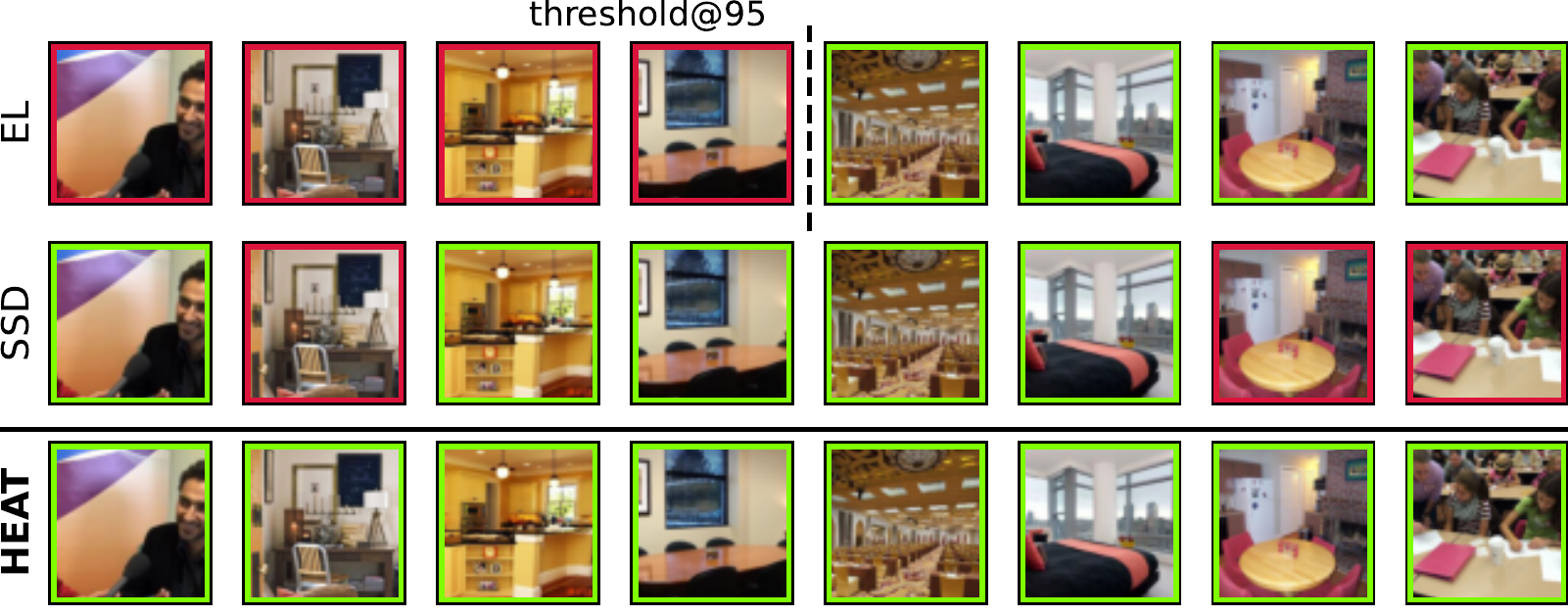}
    \caption{Qualitative comparison of {\small{HEAT}} \vs {\small{EL}}~\cite{Liu2020} and {\small{SSD}}~\cite{Sehwag2021}. Samples in green are correctly detected as OOD (LSUN), samples in red are incorrectly predicted as ID (CIFAR-10).}
    \label{fig:main_qual_results}
\end{figure*}

\newpage
\paragraph{Low data regime.} We study in~\cref{fig:number_training_samples_auc_c100} (and~\cref{sec:sup_model_analysis}) the stability of {\small HEAT} on low data regimes. Specifically, we restrict the training of {\small HEAT} to a subset of the ID dataset, \ie CIFAR-100. We compare {\small HEAT} to a fully data-driven EBM and to a {\small GMM}. The {\small EBM} is very sensitive to the lack of training data, with a gap of 12 pts {\small AUC} between 10\% of data and 100\%. On the other hand, {\small GMM} is quite robust to low data regimes with a minor gap of 0.3 pts {\small AUC} between 10\% and 100\%. {\small HEAT} builds on this stability and is able to improve the performance of {\small GMM} for all tested sampling ratios. {\small HEAT} is very stable to low data regimes which makes it easier to use than a standard {\small EBM}, it is also able to improve {\small GMM} even when few training data are available.

\subsection{Qualitative results}\label{sec:qual_results}

In~\cref{fig:main_qual_results} we display qualitative results on CIFAR-10 (ID) and show the detection results of OOD samples from LSUN. We display in red OOD samples incorrectly identified as ID samples, \ie below the threshold at 95\% of ID samples, and in green OOD samples correctly detected, \ie are above the 95\% threshold. We can see that {\small SSD} detects different OOD than {\small EL}. HEAT correctly predicts all OOD samples. Other qualitative results are provided in~\cref{sec:sup_qual_results}.

\newpage
\section{Conclusion}
\label{sec:ccl}
We have introduced the HEAT model which leverages the versatility of the {\small EBM} framework to provide a strong OOD detection method jointly effective on both far and near-OODs. HEAT i) corrects prior OOD detectors to boost their detection performances and ii) naturally combines the corrected detectors to take advantage of their strengths. 
~We  perform extensive experiments to validate HEAT on several benchmarks, highlighting the importance of the correction and the composition, and showing that HEAT sets new state-of-the-art performances on CIFAR-10, CIFAR-100, and on the large-scale Imagenet dataset. HEAT is also applicable to different backbones, and remains efficient in low-data regimes. 

Future works include extending HEAT by correcting other prior density estimators, \eg {\small KNN}. Another interesting direction is to validate HEAT on other tasks, \eg segmentation, or modalities, \eg NLP.

\section*{Acknowledgements}
This work was done under grants from the DIAMELEX ANR program (\textcolor{black}{ANR-20-CE45-0026}) and the AHEAD ANR program (\textcolor{black}{ANR-20-THIA-0002}). It was granted access to the HPC resources of IDRIS under the allocation \textcolor{black}{2021-AD011012645R1} made by GENCI.

\newpage
\clearpage
\bibliography{references}
\bibliographystyle{icml2023}

\newpage
\clearpage

\appendix
\onecolumn
\section{HEAT Method}\label{sec:sup_method}

\subsection{Energy-based models}
\label{sec:sup_ebm}
An energy-based model (EBM) is an unnormalized density model defined via its energy function $E_{\theta}:\mathbb{R}^m \rightarrow \mathbb{R}$ parameterized by a neural network with parameters $\theta$. For $\z \in \Rm$, its probability density is given by the Boltzmann distribution 

\begin{equation}
\label{supp:eq:EBM}
p_{\theta}(\z) = \frac{1}{Z_{\theta}} \exp{(\minus\Etheta(\z)}),
\end{equation}

where $Z_{\theta}$ is the partition function which is intractable in high dimension. We can train EBMs via maximum likelihood estimation:
\begin{align}
\argmax_{\theta} \; \log{p_{\theta}(\mathcal{D})}=  \argmin_{\theta} \; \mathbb{E}_{\z\sim p_{in}}[-\log{p_{\theta}(\z)}]
\end{align}
which can be approximated via stochastic gradient descent :
\begin{equation}
\theta_{i+1} = \theta_{i} - \lambda \nabla_{\theta}( \minus\log{ p_{\theta_{i}}(\z)}) \quad \text{with}\quad \z \sim p_{in}
\end{equation}

Interestingly, $\nabla_{\theta}( \minus\log{ p_{\theta_{i}}(\z)})$ can be computed without computing the intractable normalization constant $Z_{\theta}$.

We have
\begin{align*}
 \nabla_{\theta}( \minus\log{ p_{\theta}}(\z)) &= \nabla_{\theta} E_{\theta}(\z) + \nabla_{\theta} \log{ Z_{\theta}}\\
 &= \nabla_{\theta} E_{\theta}(\z) +  \frac{1}{Z_{\theta}} \nabla_{\theta} Z_{\theta} \\
 &= \nabla_{\theta} E_{\theta}(\z)  +\frac{1}{Z_{\theta}} 
\nabla_{\theta}\int_{\z} \exp(-E_{\theta}(\z)) d\z\\
 &= \nabla_{\theta} E_{\theta}(\z) + \frac{1}{Z_{\theta}} \int_{\z} 
 \nabla_{\theta}\exp(-E_{\theta}(\z)) d\z\\
 &= \nabla_{\theta} E_{\theta}(\z) + \int_{\z}
 -\nabla_{\theta}E_{\theta}(\z) \frac{\exp(-E_{\theta}(\z)) }{Z_{\theta}} d\z\\
 &= \nabla_{\theta} E_{\theta}(\z) - \E_{\z' \sim p_{\theta}} 
 [\nabla_{\theta} E_{\theta}(\z')].
\end{align*}

Therefore, training EBMs via maximum likelihood estimation (MLE) amounts to perform stochastic gradient descent with the following loss:
\begin{equation}
\label{sup:eq:CD}
\Lmle =\E_{\z \sim p_{in} }\big[\Etheta(\z)\big] - \E_{\z' \sim \ptheta}\big[\Etheta(\z')\big].
\end{equation}

Intuitively, this loss amounts to diminishing the energy for samples from the true data distribution $p(x)$ and to increasing the energy for synthesized examples sampled according from the current model. Eventually, the gradients of the energy function will be equivalent for samples from the model and the true data distribution and the loss term will be zero.
 
The expectation $\E_{\z' \sim \ptheta}\big[\Etheta(\z')\big]$ can be approximated through MCMC sampling, but we need to sample $z'$ from the model $p_{\theta}$ which is an unknown moving density. To estimate the expectation under $\ptheta$ in the right hand-side of equation (\ref{sup:eq:CD}) we must sample according to the energy-based model $\ptheta$. To generate synthesized examples from $\ptheta$, we can use gradient-based MCMC sampling such as Stochastic Gradient Langevin Dynamics (SGLD) \cite{WellingT11} or Hamiltonian Monte Carlo (HMC) \cite{Neal2011}. In this work, we use SGLD sampling following \cite{Du2019, Grathwohl2019}. In SGLD, initial features are sampled from a proposal distribution $p_0$ and are updated for $T$ steps with the following iterative rule:
\begin{equation}
\label{eq:SGLD}
\z_{t+1} = \z_{t} - \frac{\eta}{2} \nabla_{\z} E_{\theta_k}^h(\z_{t}) + \sqrt{\eta}~w_t, \;\text{~with $w_t\sim\mathcal N(\bm 0, \bm I)$}
\end{equation}
where $\eta$ is the step size. Therefore sampling from $\ptheta$ does not require to compute the normalization constant $Z_{\theta}$ either.

Many variants of this training procedure have been proposed including Contrastive Divergence (CD) \cite{Hinton2002} where $p_0 = \pdata$, or Persistent Contrastive Divergence (PCD) \cite{Tieleman2008} which uses a buffer to extend the length of the MCMC chains. We refer the reader to \cite{Song2021} for more details on EBM training with MLE as well as other alternative training strategies (score-matching, noise contrastive estimation, Stein discrepancy minimization, etc.).

\subsection{Composition function}\label{sec:sup_compo}

While many composition strategies can be consider, we choose to use the best trade-off between detection efficiency and flexibility. While combining many OOD-prior is great, hand tuning many balancing hyper-parameters can quickly become cumbersome. Our energy composition strategy $E_{\text{HEAT}}^{\beta}$ presents the advantage to have only one hyper-parameter $\beta$ to tune with a clear interpretation for its different regimes. Indeed, depending on $\beta$, this composition operator generalizes several standard aggregation operators. When $\beta \rightarrow +\infty$, we recover the maximum operator, while when $\beta \rightarrow -\infty$, we recover the minimum operator. In the $\beta \rightarrow 0$ case, we recover the sum and the resulting distribution is equivalent to a product of experts. Finally, taking  $\beta = -1$ amounts to using the \emph{logsumexp} operator which approximates a mixture of experts. In addition, to prevent the energy of one prior scorer to dominate the others, we normalize the energies using the train statistics (subtracting their mean and dividing by their standard deviation). This simple standardization gives good results in our setting, although more advanced normalization schemes could certainly be explored if needed with other prior scorers.

\section{Experiments}\label{sec:sup_exp}

\paragraph{Datasets.} We conduct experiments using CIFAR-$10$ and CIFAR-$100$ datasets \cite{krizhevsky2009learning} as in-distribution datasets. For OOD datasets, we define three categories: near-OOD datasets, mid-OOD datasets and far-OOD datasets. These correspond to different levels of proximity with the ID datasets. For CIFAR-$10$ (resp. CIFAR-$100$), we consider \texttt{TinyImagenet}\footnote{The dataset can be found at: \url{https://www.kaggle.com/c/tiny-imagenet}} and CIFAR-$100$ (resp. CIFAR-$10$) as near-OOD datasets. Then for both CIFAR-$10$ and CIFAR-$100$, we use \texttt{LSUN} \cite{yu15lsun} and \texttt{Places} \cite{zhou2017places} datasets as mid-OOD datasets, and \texttt{Textures} \cite{cimpoi14describing} and \texttt{SVHN} \cite{svhn-dataset} as far-OOD datasets. We use different Imagenet~\cite{deng2009imagenet} benchmarks. In~\cref{tab:ood_results_imagenet} we use the benchmarks of~\cite{Sehwag2021,Sun2022} with \texttt{iNaturalist}~\cite{van2018inaturalist},  \texttt{LSUN}~\cite{yu15lsun}, \texttt{Places}~\cite{zhou2017places} and \texttt{Textures}~\cite{cimpoi14describing}. In~\cref{sec:sup_openood} we use the Imagent benchmark recently introduced in OpenOOD~\cite{OpenOOD2022}, and we refer the reader to the paper for details about the dataset\footnote{Datasets for the OpenOOD benchmark can be downloaded using: \url{https://github.com/Jingkang50/OpenOOD}.}. Finally in~\cref{sec:sup_vit} we ue the Imagenet benchmark proposed in~\cite{Wang2022}, with notably the OpenImage-O dataset introduced specifically as an OOD dataset for the Imagenet benchmark in~\cite{Wang2022}, for more details we refer the reader to the paper.

\paragraph{Implementation details.} All experiments were conducted using \texttt{PyTorch} \cite{NEURIPS2019_9015}. We use a ResNet-34 classifier from the \texttt{timm} library \cite{rw2019timm} for the CIFAR-10 and CIFAR-100 datasets and a ResNet-50 for the Imagenet experiments. HEAT consists in a $6$ layers MLP trained for $20$ epochs with Adam with learning rate $5e\text{-}6$. The network input dimension is $512$ (which is the dimension of the penultimate layer of ResNet-34) for the CIFAR-10/100 benchmarks and $2048$ (which is the dimension of the penultimate layer of ResNet-50) for the Imagenet benchmark. The hidden dimension is $1024$ for CIFAR-10/100 and $2048$ for Imagenet, and the output dimension is $1$. For SGLD sampling, we use 20 steps with an initial step size of $1e\text{-}4$ linearly decayed to $1e\text{-}5$ and an initial noise scale of $5e\text{-}3$ linearly decayed to $5e\text{-}4$. We add a small Gaussian noise with std $1e\text{-}4$ to each input of the EBM network to stabilize training as done in previous works \cite{Du2019, Grathwohl2019}. The $L_2$ coefficient is set to $10$. We use temperature scaling on the mixture of Gaussian distributions energy with temperature $T_{\cG} = 1e3$. The hyper-parameters for the CIFAR-$10$ and CIFAR-$100$ models are identical.

\paragraph{Additional metric} In~\cref{sec:sup_openood} we use an additional metric the {\small AUPR}, which  measures the area under the Precision-Recall (PR) curve, using the ID samples as positives (see~\cite{OpenOOD2022} for details). The score corresponds to the {\small AUPR}-In metric in other works.

\subsection{Additional comparison to state-of-the-art}\label{sec:sup_sota}

\subsubsection{Imagenet OpenOOD results}\label{sec:sup_openood}

We compare HEAT on the recently introduced OpenOOD benchmark~\cite{OpenOOD2022} in~\cref{tab:imagenet_openood_aggregated,tab:imagenet_openood_farood,tab:ood_results_imagenet_openood}. In addition to the baselines used in the main paper, the OpenOOD benchmark includes the Mahalanobis detector \cite{mahalanobis2018} (MDS), OpenMax \cite{BendaleB16}, the Gram matrix detector \cite{Sastry2020} (Gram), KL matching \cite{HendrycksBMZKMS22} (KLM) and GradNorm \cite{HuangGL21}. We show that on the aggregated results in~\cref{tab:imagenet_openood_aggregated} HEAT outperforms previous methods, and sets new state-of-the-art performances for our considered setting. Similarly to our comparison in Sec. 4.2 we can see that HEAT performs well on far-OOD~\cref{tab:imagenet_openood_farood} and near-OOD~\cref{tab:ood_results_imagenet_openood}. On far-OOD HEAT has the best performances on each metric, \ie 2.4 {\small FPR95}, 99.4 {\small AUC} and 100 {\small AUPR}. On near-OOD HEAT has the best performances on {\small AUC}, \ie +2.6 pts {\small AUC} \vs KNN, -0.8 pts {\small FPR95} and +0.9 pts {\small AUPR} \vs ReAct.  

\begin{table*}[ht]
\begin{minipage}[c]{0.49\linewidth}
\centering
\caption{\textbf{Aggregated results on the Imagenet OpenOOD benchmark}. All methods are based on an Imagnet pre-trained ResNet-50.}
\begin{subtable}[c]{\textwidth}
    \centering
	   \resizebox{1\linewidth}{!}{%
    	\begin{tabular}{l cc|c}
    		\toprule
  	 	       \multirow{2}{*}{Method} & \textbf{Near-OOD}  & \textbf{Far-OOD} & \textbf{Average}\\
        &  \tiny{\text{FPR95$\downarrow$ / AUC $\uparrow$ / AUPR $\uparrow$}} 
        & \tiny{\text{FPR95$\downarrow$ / AUC $\uparrow$ / AUPR $\uparrow$}} & \tiny{\text{FPR95$\downarrow$ / AUC $\uparrow$ / AUPR $\uparrow$}}  \\
    		\midrule 
                OpenMax & 76.3 / 66.0 / 92.2 & 55.0 / 84.9 / 97.0 & 69.2 / 72.3 / 93.8 \\
                MSP & 73.7 / 69.3 / 94.6 & 57.8 / 86.2 / 97.5 & 68.4 / 74.9 / 95.6 \\
                ODIN &	68.2 / 73.2 / 95.0	& 21.8 / 94.4 / 99.1 & 52.7 / 80.2 / 96.3 \\
                MDS & 86.9 / 68.3 / 90.8 & 18.4 / 94.0 / 98.5 & 64.1 / 76.8 / 93.3\\
                Gram &	83.1 / 68.3 / 91.8	& 43.2 / 89.2 / 97.9 & 69.8 / 75.3 / 93.8\\
                EL & 73.3 / 73.5 / 95.3 & 33.8 / 92.8 / 98.8 & 60.1 / 79.9 / 96.4\\
                GradNorm & 61.1 / 75.7 / 94.9 & {15.0} / 95.8 / \underline{99.3} & \underline{45.7} / 82.4 / 96.3\\
                ReAct & \underline{57.2} / {79.3} / \underline{96.2}	& 23.8 / 95.2 / 99.3 & 46.1 / 84.6 / \underline{97.2} \\
                MLS & 72.2 / 73.6 / 95.3 & 37.6 / 92.3 / 98.7 & 60.7 / 79.8 / 96.5 \\
                KLM & 68.1 / 73.4 / 94.8 & 56.6 / 88.8 / 98.1 & 64.3 / 78.5 / 95.9 \\
                VIM & 73.8 / 79.9 / 95.7 & \underline{6.9} / 98.4 / \textbf{99.8} & 51.5 / 86.1 / 97.1\\
                KNN  & 71.9 / \underline{80.8} / \underline{95.7} & 8.4 / 98.0 / \textbf{99.7} & 50.8 / \underline{86.5} / 97.0\\
                DICE  & 65.1 / 73.8 / 95.1 & 15.8 / 95.7 / 99.3 & 48.6 / 81.1 / 96.5 \\
                \midrule
        \rowcolor[gray]{.95} \textbf{HEAT} & \textbf{55.2} / \textbf{84.8} / \textbf{97.1} & \textbf{2.6} / \textbf{99.4} / \textbf{100.0} & \textbf{37.7} / \textbf{89.6} / \textbf{98.1} \\
    		\bottomrule
    	\end{tabular}
    }
\end{subtable}
\label{tab:imagenet_openood_aggregated}
\end{minipage}%
\hspace{1em}
\begin{minipage}[c]{0.49\linewidth}
\centering
\caption{\textbf{Results on far-OOD of the Imagenet OpenOOD benchmark}. All methods are based on an Imagnet pre-trained ResNet-50.}
\begin{subtable}[c]{\textwidth}
    \centering
	   \resizebox{1\linewidth}{!}{%
    	\begin{tabular}{l cc|c}
    		\toprule
  	 	       \textbf{Method} & \textbf{Textures} & \textbf{MNIST} & \textbf{Far-OOD} \\
        & \tiny{\text{FPR95$\downarrow$ / AUC $\uparrow$ / AUPR $\uparrow$}} & \tiny{\text{FPR95$\downarrow$ / AUC $\uparrow$ / AUPR $\uparrow$}} & \tiny{\text{FPR95$\downarrow$ / AUC $\uparrow$ / AUPR $\uparrow$}} \\
    		\midrule 
                OpenMax & 65.3 / 78.9 / 96.0	& 44.6 / 90.9 / 98.0	& 55.0 / 84.9 / 97.0 \\
                MSP & 63.6 / 82.5 / 97.2	& 52.0 / 89.8 / 97.8	& 57.8 / 86.2 / 97.5 \\
                ODIN &	42.5 / 89.2 / 98.3	& \textbf{0.9} / \textbf{99.7 }/ \textbf{99.9}	& 21.8 / 94.4 / 99.1 \\
                MDS & 36.7 / 90.2 / 97.4	& \textbf{0.0} / 97.7 / \textbf{99.6}	& 18.4 / 94.0 / 98.5 \\
                Gram &	53.3 / 82.7 / 96.7	& 33.1 / 95.7 / 99.2	& 43.2 / 89.2 / 97.9 \\
                EL &	49.2 / 88.8 / 98.3	& 18.3 / 96.8 / 99.4	& 33.8 / 92.8 / 98.8 \\
                GradNorm &	29.4 / 92.1 / 98.7	& \textbf{0.6 / 99.5 / 99.9}	& {15.0} / 95.8 / \underline{99.3} \\
                ReAct &	43.1 / 91.6 / 98.9	& 4.5 / 98.8 / 99.8	& 23.8 / 95.2 / 99.3  \\
                MLS & 51.3 / 88.5 / 98.3	& 24.0 / 96.1 / 99.2	& 37.6 / 92.3 / 98.7  \\
                KLM & 67.6 / 84.7 / 97.6	& 45.5 / 92.9 / 98.6	& 56.6 / 88.8 / 98.1  \\
                VIM &	\underline{12.4} / \underline{97.5} / \textbf{99.7}	& \underline{1.4} / \underline{99.2} / \textbf{99.9}	& \underline{6.9} / 98.4 / \textbf{99.8} \\
                KNN & {16.9} / {96.2} / \textbf{99.5}	& \textbf{0.0} / \textbf{99.9} / \textbf{99.9}	& 8.4 / 98.0 / \textbf{99.7} \\
                DICE & 29.4 / 92.1 / 98.7	& \underline{2.3} / \underline{99.3} / \textbf{99.9}	& 15.8 / 95.7 / 99.3  \\
                \midrule
        \rowcolor[gray]{.95} \textbf{HEAT} & \textbf{5.3 }/\textbf{ 98.9} / \textbf{99.9}&  \textbf{0.0} / \textbf{99.9} / \textbf{100.0}& \textbf{2.6} / \textbf{99.4} / \textbf{100.0} \\
    		\bottomrule
    	\end{tabular}
    }
\end{subtable}
\label{tab:imagenet_openood_farood}
\end{minipage}%
\end{table*}

\begin{table*}[ht]
\caption{\textbf{Results on near-OOD of the Imagenet OpenOOD benchmark}. All methods are based on an Imagnet pre-trained ResNet-50.}
\setlength\tabcolsep{10pt}
\begin{subtable}[c]{\textwidth}
    \centering
	   \resizebox{\linewidth}{!}{%
    	\begin{tabular}{l cccc|c}
    		\toprule
  	 	       \textbf{Method} & \textbf{Species} & \textbf{iNaturalist} & \textbf{OpenImage-O} & \textbf{Imagenet-O} & \textbf{Near-OOD} \\
        &  \scriptsize{\text{FPR95$\downarrow$ / AUC $\uparrow$ / AUPR $\uparrow$}} & \scriptsize{\text{FPR95$\downarrow$ / AUC $\uparrow$ / AUPR $\uparrow$}} & \scriptsize{\text{FPR95$\downarrow$ / AUC $\uparrow$ / AUPR $\uparrow$}} & \scriptsize{\text{FPR95$\downarrow$ / AUC $\uparrow$ / AUPR $\uparrow$}} & \scriptsize{\text{FPR95$\downarrow$ / AUC $\uparrow$ / AUPR $\uparrow$}} \\
    		\midrule 
                OpenMax &	81.8 / 70.8 / 90.5 &	56.8 / 79.5 / 92.7 &	66.7 / 81.5 / 91.4 & 99.9 / 32.1 / 94.1 & 76.3 / 66.0 / 92.2 \\
                MSP &	79.2 / 75.2 / \underline{92.9} &	52.4 / 88.5 / 97.1 &	63.2 / 85.0 / 94.2 & 100.0 / 28.7 / 94.3 & 73.7 / 69.3 / 94.6 \\
                ODIN &	80.5 / 71.6 / 91.4 &	42.0 / 91.2 / 97.7 &	50.5 / 88.4 / 95.3 & 99.9 / 41.5 / 95.4 & 68.2 / 73.2 / 95.0 \\
                MDS &	94.8 / 60.2 / 87.6 &	93.7 / 67.8 / 90.9 &	82.4 / 70.7 / 86.4 & \textbf{76.9} / \textbf{74.3} / \textbf{98.2} & 86.9 / 68.3 / 90.8 \\
                Gram &	86.8 / 67.3 / 89.9 &	75.2 / 78.4 / 94.1 &	79.0 / 71.9 / 86.8 & 91.5 / 55.8 / 96.4 & 83.1 / 68.3 / 91.8	\\
                EL &	82.3 / 72.1 / 91.6 &	53.7 / 90.6 / 97.8 &	57.0 / 89.2 / 96.0 & 100.0 / 42.0 / 95.7 & 73.3 / 73.5 / 95.3	\\
                GradNorm &	\underline{74.4} / 75.7 / 92.8 &	\underline{26.9} / \underline{93.9} / \underline{98.4} &	47.5 / 85.2 / 92.8 & 95.6 / 48.2 / 95.5 & 61.1 / 75.7 / 94.9	\\
                ReAct &	\textbf{68.4} / \textbf{77.5} / \textbf{92.6} &	\textbf{19.3} / \textbf{96.4 }/ \textbf{99.2} &	\underline{43.5} / \underline{90.6} / \underline{96.4} & 98.1 / 52.5 / 96.6 & \underline{57.2} / {79.3} / \underline{96.2} \\
                MLS &	80.8 / 73.0 / 91.8 &	50.8 / 91.2 / 97.9 &	57.1 / 89.3 / 96.0 & 100.0 / 41.0 / 95.6 & 72.2 / 73.6 / 95.3 \\
                KLM &	\textbf{73.7} / 74.5 / 91.7 &	41.1 / 90.8 / 97.4 &	57.8 / 87.4 / 94.7 & 100.0 / 40.8 / 95.4 & 68.1 / 73.4 / 94.8 \\
                VIM &	84.0 / 70.7 / 90.9 &	68.0 / 88.4 / 97.4 &	57.7 / 89.6 / 96.4 & 85.5 / 70.9 /\textbf{ 98.3} & 73.8 / 79.9 / 95.7 \\
                KNN &	76.4 / \underline{76.4} / \textbf{93.0} &	68.6 / 85.0 / 96.4 &	58.0 / 86.4 / 94.9 & 84.8 / 75.4 / \textbf{98.6} & 71.9 / \underline{80.8} / \underline{95.7} \\
                DICE &	78.6 / 71.3 / 91.3 &	35.3 / 92.5 / 98.1 &	\underline{47.7} / 88.5 / 95.3 & 98.5 / 42.9 / 95.5 & 65.1 / 73.8 / 95.1 \\
                \midrule
        \rowcolor[gray]{.95} \textbf{HEAT} & 74.3 / 76.8 / \textbf{93.7}& 27.4 / 95.0 / 98.9 &  \textbf{41.4} / \textbf{91.8} / \textbf{97.2}&  77.6 / 75.5 / \textbf{98.7}& \textbf{55.2} / \textbf{84.8} / \textbf{97.1} \\
    		\bottomrule
    	\end{tabular}
    }
\end{subtable}
\label{tab:ood_results_imagenet_openood}
\end{table*}

\newpage
\clearpage
\subsubsection{ViT results}\label{sec:sup_vit}

In~\cref{tab:ood_results_imagenet_vim_vit} we compare HEAT using a Vision Transformer\footnote{The model used can be found at \url{https://github.com/haoqiwang/vim}} (ViT), on the Imagenet benchmark introduced in~\cite{Wang2022}. We show that on the aggregated results HEAT outperforms the previous best method, ViM~\cite{Wang2022}, by -1.7 pts {\small FPR95}. Importantly HEAT ouperforms other method on three datasets of the benchmark, \ie OpenImage-O, Textures, Imagenet-O, and is competitive on iNaturalist. \cref{tab:ood_results_imagenet_vim_vit} demonstrates the ability of HEAT to adapt to architectures of neural networks, \ie Vision Transformer~\cite{dosovitskiy2020image}, other than the convolutional networks (\ie ResNet-34 \& ResNet-50) tested in~\cref{sec:sota_comparison}.

\begin{table}[!ht]
\caption{\textbf{Results on Imagenet.} All methods are based on an Imagenet pre-trained \textbf{Vision Transformer} (ViT) model. $\uparrow$ indicates larger is better and $\downarrow$ the opposite. }
\setlength\tabcolsep{12pt}
    \centering
	   \resizebox{0.9\linewidth}{!}{%
    	\begin{tabular}{l  c c  c c | c}
    		\toprule
  	 	     \multirow{2}{*}{\textbf{Method}} & \textbf{OpenImage-O} & \textbf{Textures} & \textbf{iNaturalist} & \textbf{Imagenet-O} & \textbf{Average}\\
      &   \tiny{\text{FPR95$\downarrow$ / AUC $\uparrow$}} &\tiny{\text{FPR95$\downarrow$ / AUC $\uparrow$}} & \tiny{\text{FPR95$\downarrow$ / AUC $\uparrow$}} & \tiny{\text{FPR95$\downarrow$ / AUC $\uparrow$}} & \tiny{\text{FPR95$\downarrow$ / AUC $\uparrow$}} \\
    		\midrule
            MSP & 34.2 / 92.5 & 48.6 / 87.1 & 19.0 / 96.1 & 64.8 / 81.9 & 41.7 / 89.4 \\
            EL & 14.0 / 97.1 & 28.2 / 93.4 & 6.2 / 98.7 & 41.3 / 90.5 & 22.4 / 94.9 \\
            ODIN & 15.7 / 96.9 & 30.6 / 93.0 & 6.6 / 98.6 & 44.2 / 89.9 & 24.3 / 94.6 \\
            MaxLogit & 15.7 / 96.9 & 30.6 / 93.0 & 6.6 / 98.6 & 44.2 / 89.9 & 24.3 / 94.6 \\
            KL Matching & 28.5 / 93.9 & 44.1 / 88.8 & 14.8 / 96.9 & 55.7 / 84.1 & 35.8 / 90.9 \\
            KNN & 45.8 / 91.7&  28.9 / 93.2&  52.3 / 91.1&  52.9 / 88.4&   45.0 / 91.1\\
            Residual & 32.6 / 92.7 & 33.8 / 92.2 & 6.6 / 98.6 & 47.9 / 88.2 & 30.2 / 92.9 \\
            ReAct & 13.5 / 97.4 & 28.5 / 93.3 & 4.3 / 99.0 & 42.6 / 90.7 & 22.2 / 95.1 \\
            Mahalanobis & 13.5 / 97.5 & 25.2 / 94.2 & \textbf{2.1} / \textbf{99.5} & 37.0 / \underline{92.8} & 19.5 / 96.0 \\
            ViM & \underline{12.6} / \underline{97.6} & \underline{20.3} / \underline{95.3} & \underline{2.6} / \textbf{99.4} & \underline{36.8} / 92.6 & \underline{18.1} / \underline{96.2} \\
                \midrule
              \rowcolor[gray]{.95} \textbf{HEAT} & \textbf{11.2} / \textbf{97.8}&  \textbf{12.8} / \textbf{96.9}&  6.9 / 98.2&  \textbf{34.8} / \textbf{93.1}&   \textbf{16.4} / \textbf{96.5} \\
        		\bottomrule
    	\end{tabular}
    }
\label{tab:ood_results_imagenet_vim_vit}
\end{table}

\subsubsection{Results Imagenet with DICE}\label{sec:sup_dice}

In this section we compare the performance of HEAT when using DICE instead of EL as one its components. We show in \cref{tab:ood_results_imagenet_dice} that using DICE instead of EL as part of HEAT's components further improves the OOD detection performances on all OOD datasets except Textures. 

\begin{table*}[!ht]
\caption{\textbf{Results of HEAT using DICE vs EL on Imagenet.} All methods use an Imagenet pre-trained {ResNet-50}. Results are reported with \text{{\small FPR95}$\downarrow$ / {\small AUC} $\uparrow$}.}
\setlength\tabcolsep{10pt}
\centering
\begin{tabular}{l  c c  c c | c}
    \toprule
    \textbf{Method} & \textbf{iNaturalist} & \textbf{SUN} & \textbf{Places} & \textbf{Textures} & \textbf{Average}\\
    \midrule
    \textbf{HEAT} (w/. EL) & {28.1} / {94.9}&  {44.6} / {90.7} &  {58.8} / {86.3}&  {\textbf{5.9} / \textbf{98.7}}&   {{34.4} / {92.6}} \\[1pt]
     \textbf{HEAT} (w/. DICE) & \textbf{24.4 / 95.4}& \textbf{39.5} / \textbf{91.4}&  \textbf{53.2} / \textbf{87.4}&  {9.7} / {97.5}&   \textbf{31.7 / 93.1} \\
    \bottomrule
\end{tabular}
\label{tab:ood_results_imagenet_dice}
\end{table*}

\subsubsection{Results Supervised Contrastive backbone}\label{sec:sup_supcon}

In this section we evaluate HEAT when using a supervised contrastive backbone and compare with KNN+ and SSD+. We show in \cref{tab:ood_results_imagenet_supcon} that HEAT still largely outperforms the competition even with the supervised contrastive backbone.

\begin{table}[!ht]
\caption{\textbf{Results on Imagenet.} All methods use an Imagenet Supervised Contrastive pre-trained {ResNet-50}. Results are reported with \text{{\small FPR95}$\downarrow$ / {\small AUC} $\uparrow$}.}
\setlength\tabcolsep{10pt}
\centering
\resizebox{0.7\linewidth}{!}{%
\begin{tabular}{l  c c  c c | c}
    \toprule
    \textbf{Method} & \textbf{iNaturalist} & \textbf{SUN} & \textbf{Places} & \textbf{Textures} & \textbf{Average}\\
    \midrule
    SSD+  & 34.3 / 95.0&  65.2 / 86.3&  70.2 / 83.8&  14.7 / 95.6&   46.1 / 90.2 \\
    KNN+ & 30.8 / 94.7 & 48.9 / 88.4 & 60.0 / 84.6 & 17.0 / 94.5 & 39.2 / 90.6 \\
    \midrule
    \rowcolor[gray]{.95} \textbf{HEAT} & \textbf{14.6 / 97.2}&  \textbf{32.4 / 93.4}&  \textbf{45.4 / 89.9}&  \textbf{9.7 / 97.7}&   \textbf{25.5 / 94.6} \\
    \bottomrule
\end{tabular}
}
\label{tab:ood_results_imagenet_supcon}
\end{table}

\subsection{Model analysis}\label{sec:sup_model_analysis}

\begin{figure*}[ht]
\vspace{12pt}
\begin{minipage}[t]{0.48\linewidth}
\vspace{0pt}
\begin{minipage}[t]{0.49\linewidth}
    \vspace{0pt}
        \centering
    \includegraphics[width=1\linewidth]{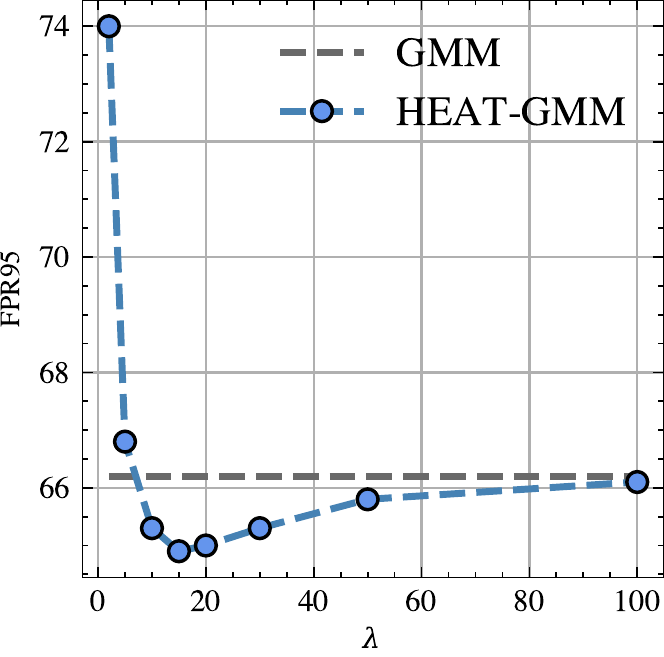}
    \subcaption{$\lambda$ \vs {\small FPR95}$\downarrow$}
    \label{fig:lambda_analysis_auc_c100}
\end{minipage}
\begin{minipage}[t]{0.49\linewidth}
\vspace{0pt}
        \centering
    \includegraphics[width=1\linewidth]{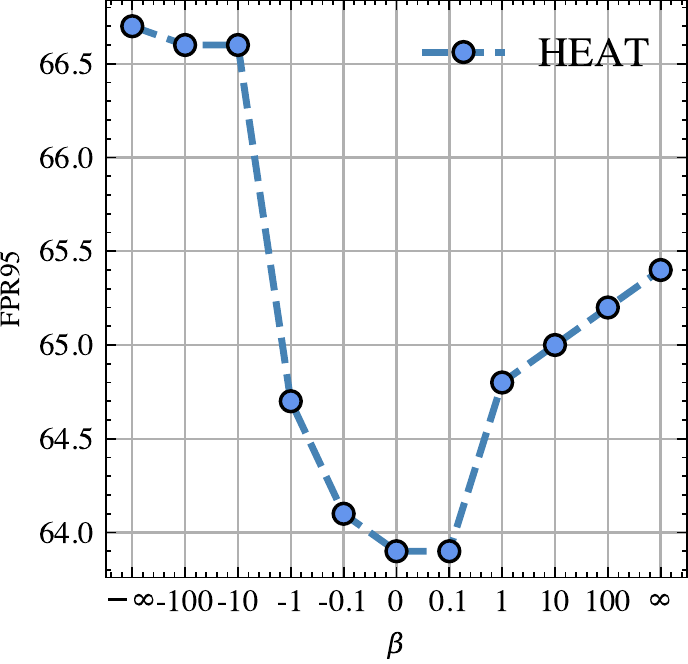}
   \subcaption{$\beta$ \vs {\small FPR95}$\downarrow$}
   \label{fig:beta_analysis_fpr_c100}
\end{minipage}
\caption{On CIFAR-100 ID: (a) impact of $\lambda$ in~\cref{eq:heat_training_objective} \vs ~{\small FPR95} and (b) analysis of $\beta$ in~\cref{eq:fusion_function} \vs ~{\small FPR95}. }
\label{fig:lambda_beta_analysis_c100}
\end{minipage}
\hfill
\begin{minipage}[t]{0.48\linewidth}
\vspace{0pt}
    \centering
    \includegraphics[width=0.93\linewidth]{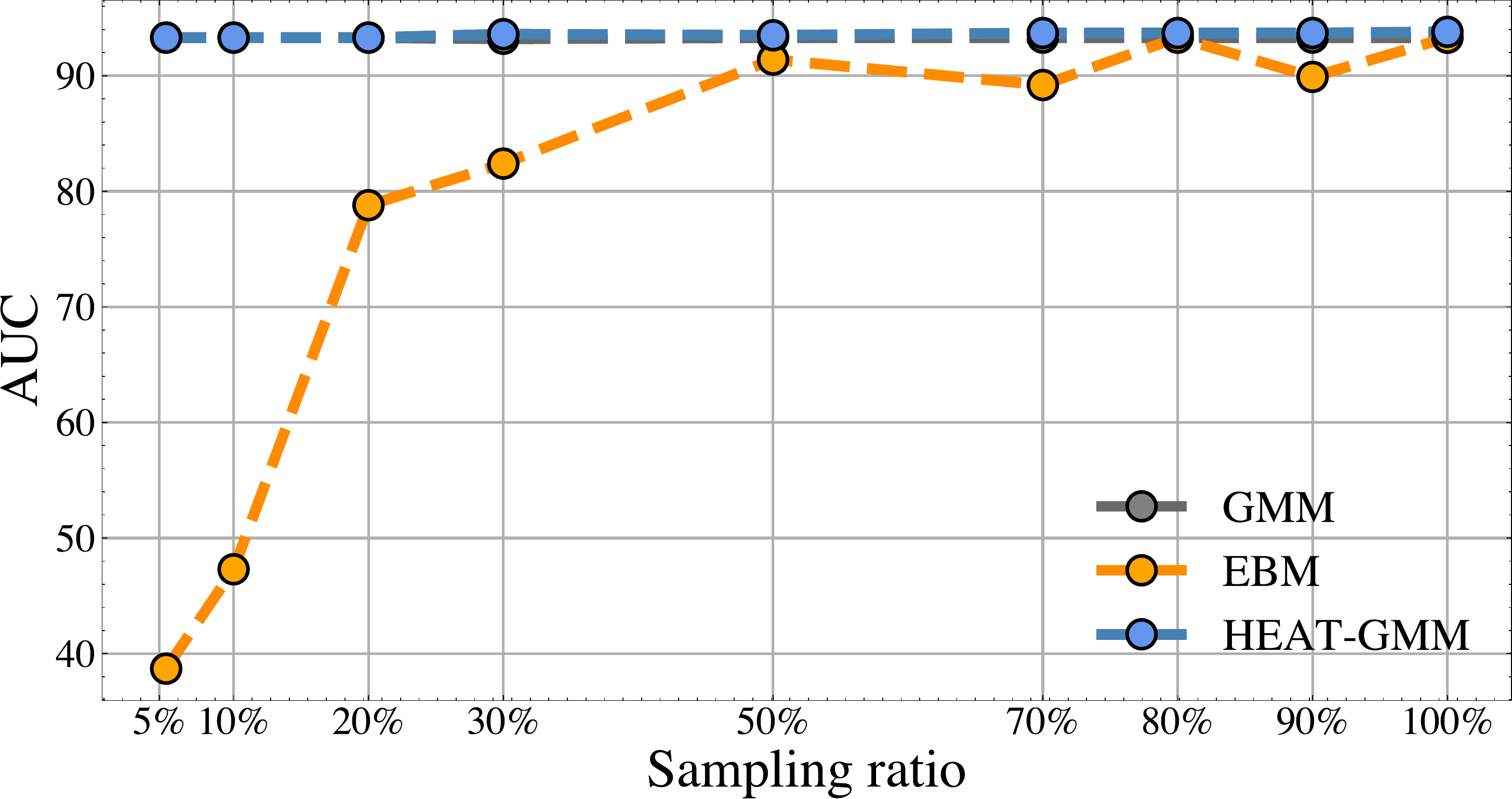}
    \caption{Impact on performances ({\small AUC}$\uparrow$ on CIFAR-10) \vs the number of training data for {\small GMM} density, fully data-driven {\small EBM}, and HEAT. Our hybrid  approach maintains strong performances in low-data regime, in contrast to the fully data-driven {\small EBM}.}
\label{fig:number_training_samples_auc_c10}
\end{minipage}
\end{figure*}

In~\cref{fig:lambda_beta_analysis_c100} we show the impact of $\lambda$ in~\cref{eq:heat_training_objective} and $\beta$ \vs {\small FPR95} on CIFAR-100, we study in~\cref{fig:number_training_samples_auc_c10} how HEAT behaves on low data regimes with CIFAR-10 as ID dataset. Finally in~\cref{tab:compute_cost} we study the computational requirements of HEAT.

\paragraph{Robustness to $\bm \lambda$} In~\cref{fig:lambda_analysis_auc_c100} we can see that we have similar trends to~\cref{fig:lambda_analysis_auc_c10}. For values of $\lambda$ too high, \ie when the expressivity of the energy-based correction is limited, {\small HEAT-GMM} has the same performances than {\small GMM}. For values of $\lambda$ too low the energy-based correction is not controlled and disregards the prior scorer, \ie {\small GMM}. Finally for a wide range of $\lambda$ values {\small HEAT-GMM} improves the OOD detection performances of {\small GMM}.

\paragraph{Robustness to $\bm \beta$} In~\cref{fig:beta_analysis_fpr_c100} we show that HEAT is stable wrt. $\beta$ on CIFAR-100 similarly to~\cref{fig:beta_analysis_fpr_c10}.

\paragraph{Low  data regime} Similarly to~\cref{fig:number_training_samples_auc_c100}, we can see that training solely an {\small EBM} is very unstable when the number of data is low. On the other hand {\small HEAT-GMM} is stable to the lack of data and improves {\small GMM} even with few ID samples available.

\paragraph{Computational cost} In~\cref{tab:compute_cost} we report the cost of computing of different components of HEAT, \eg forward pass of a ResNet-50, energy computation of {\small GMM}. We extrapolate based on those inference time the computational cost of deep-ensembles~\cite{deepensembles2017} and of HEAT. The compute time for HEAT, 5.194ms, is due at 84\% by the inference time of a ResNet-50. This is why deep-ensembles has a compute time of 2500ms which is 4.8 times larger than that of HEAT. Further more we can see that correcting {\small GMM} with HEAT only brings an overhead of 1ms, which will not scale with the size of the model but only its embedding size, \eg CLIP~\cite{Radford2021} as an embedding size of 1024 for its largest model.

\begin{table}[!ht]
\caption{Computational cost reported in \textbf{ms} $\downarrow$. Times are reported using RGB images of size $224\times224$, a ResNet-50 with an output size of 2048, 1000 classes (\ie Imagenet setup), on a single GPU (Quadro RTX 6000 with 24576MiB).}
\setlength\tabcolsep{12pt}
    \centering
	   \resizebox{0.8\linewidth}{!}{%
    	\begin{tabular}{c  c c  c c | c c }
    		\toprule
  	 	     ResNet-50  & GMM & GMM-std & EL & EBM  & deep-ensemble & \textbf{HEAT}  \\
    		\midrule
                500 & 8 & 8 & 0.4 & 1 & 2501 & 519.4 \\
        		\bottomrule
    	\end{tabular}
    }
\label{tab:compute_cost}
\end{table}

\newpage
\clearpage

\subsection{Qualitative results}\label{sec:sup_qual_results}

\begin{figure}[h]
    \centering
    \includegraphics[width=\linewidth]{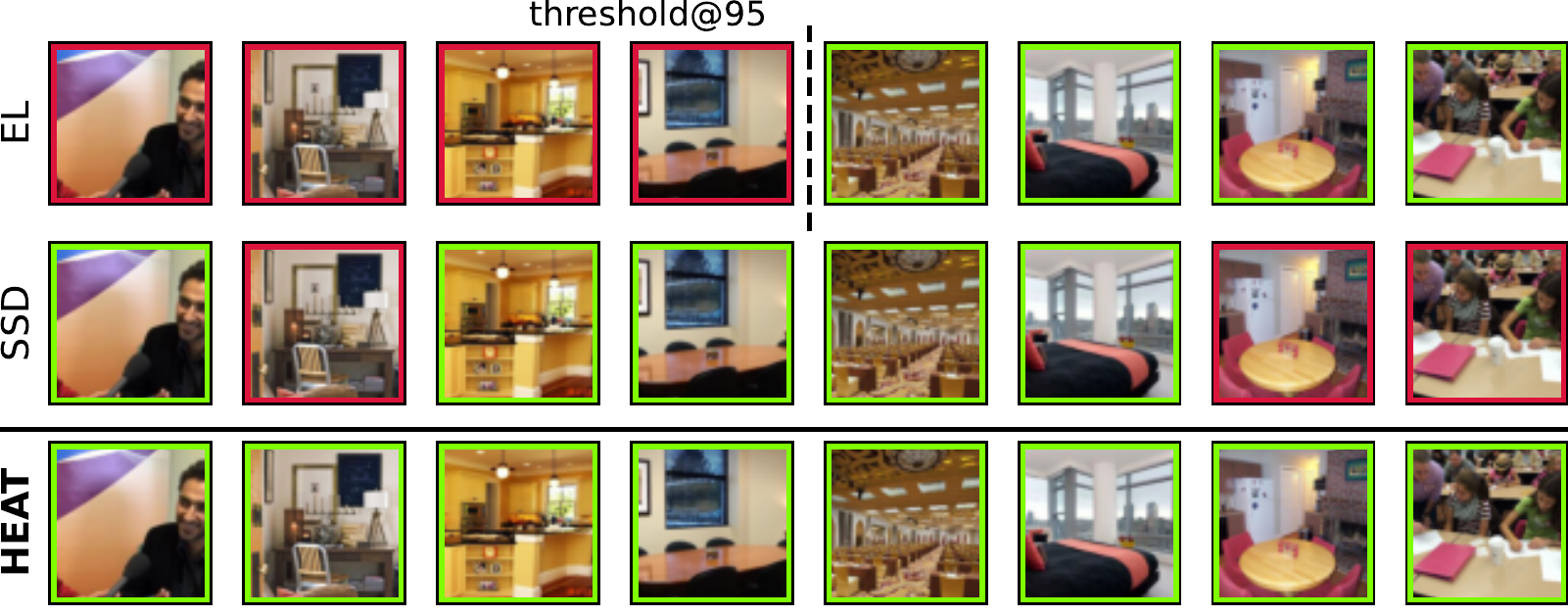}
    \caption{Qualitative comparison of HEAT \vs EL~\cite{Liu2020} and SSD~\cite{Sehwag2021} on \textbf{LSUN}. Samples in green are correctly detected as OOD (above the 95\% of ID threshold), samples in red are incorrectly predicted as ID, \ie an energy lower than the threshold.}
    \label{fig:sup_lsun_qual_results}
\end{figure}

\begin{figure}[h]
    \centering
    \includegraphics[width=\linewidth]{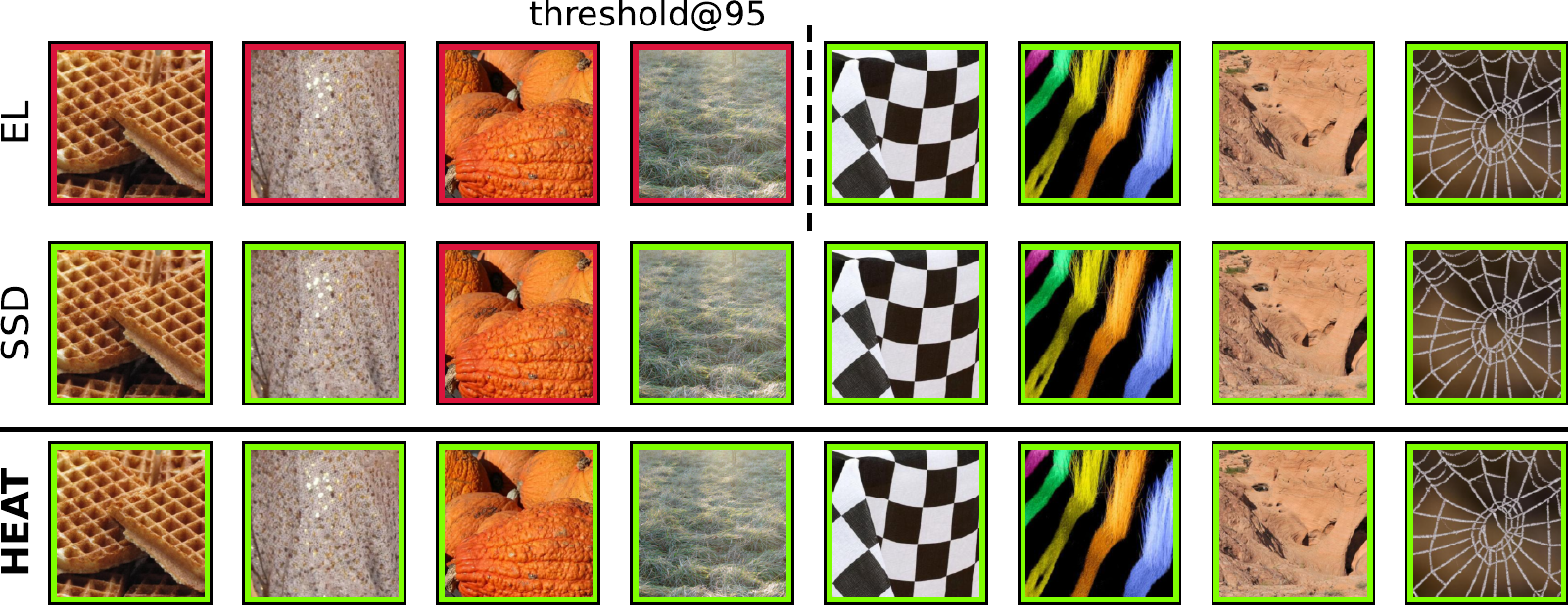}
    \caption{Qualitative comparison of HEAT \vs EL~\cite{Liu2020} and SSD~\cite{Sehwag2021} on \textbf{Textures}. Samples in green are correctly detected as OOD (above the 95\% of ID threshold), samples in red are incorrectly predicted as ID, \ie an energy lower than the threshold.}
    \label{fig:sup_textures_qual_results}
\end{figure}

We show qualitative results of HEAT \vs EL~\cite{Liu2020} and SSD~\cite{Sehwag2021} on LSUN~\cref{fig:sup_lsun_qual_results} and Textures~\cref{fig:sup_textures_qual_results}. On~\cref{fig:sup_lsun_qual_results} we can see that EL and SSD detect different OOD samples. HEAT is able though the correction and composition to recover those mis-detected OOD samples. On~\cref{fig:sup_textures_qual_results} we can see that SSD performs well on the far-OOD dataset (Textures), however HEAT is able to recover a mis-detected OOD sample. \cref{fig:sup_lsun_qual_results} and \cref{fig:sup_textures_qual_results} qualitatively show how HEAT is able to better mis-detect OOD samples.

\end{document}